\documentclass[10pt,twocolumn,letterpaper]{article}

\usepackage{cvpr}              
\definecolor{cvprblue}{rgb}{0.21,0.49,0.74}
\usepackage[pagebackref,breaklinks,colorlinks,allcolors=cvprblue]{hyperref}

\def\paperID{34280} 
\def\confName{CVPR}
\def\confYear{2026}

\title{Learning to Adapt: Self-Improving Web Agent via Cognitive-Aware Exploration}

\author{Weile Chen$^{1}$\footnotemark[2] \quad Bingchen Miao$^{1}$\footnotemark[2]\quad  Qifan Yu$^1$\footnotemark[2] \quad Wendong Bu$^1$ \quad Guoming Wang$^{1}$\footnotemark[1] \quad 
Wenqiao Zhang$^{1}$\footnotemark[1] \\
Shengyu Zhang$^1$\quad\quad Juncheng Li$^{1}$\footnotemark[1]\quad\quad Siliang Tang$^1$\\ 
\small $^1$Zhejiang University \\
{\tt\small Corresponding authors:\{Nb21013,wenqiaozhang,junchengli\}@zju.edu.cn}
}

\usepackage{algorithm}
\usepackage{algorithmic}
\usepackage{multirow}

\usepackage{amsmath}

\usepackage{booktabs}
\usepackage{calc}
\usepackage{subcaption}
\usepackage[most]{tcolorbox}
\usepackage{enumitem}

\begin{document}
\maketitle

\renewcommand{\thefootnote}{\fnsymbol{footnote}}
\footnotetext[2]{Equal Contribution.}
\footnotetext[1]{Corresponding authors.}

\begin{abstract}
Recent advances in Multimodal Large Language Models (MLLMs) have led to promising progress in web agents. However, existing web agents often rely on handcrafted execution pipelines or expensive expert trajectories, limiting their adaptability to complex, dynamic environments. To address these challenges, we propose \textbf{SCALE} (\textbf{S}elf-\textbf{C}ognitive-\textbf{A}ware \textbf{L}earning and \textbf{E}xploration), which leverages three adversarial roles---\textit{Selector}, \textit{Predictor}, and \textit{Judger} to autonomously discover the agent’s limitations and expand its cognitive boundaries through environmental exploration. Moreover, we propose \textbf{SCALE-Hop}, a graph exploration strategy that facilitates global planning and helps agents avoid local exploration traps. To further support learning, we construct \textbf{SCALE-20k}, a large-scale dataset collected from 19 real-world websites, containing diverse task types and structured demonstrations generated from SCALE’s exploration traces. Experimental results show that our approach significantly improves the performance and generalization of multiple MLLMs in various web environments. Our framework offers a scalable and generalizable solution for building truly autonomous and adaptive web agents.

\end{abstract}  
\section{Introduction}
\label{sec:intro}

Recently, with the development of Multimodal Large Language Models (MLLMs), MLLM-based web agents~ \cite{lin2025showui,zheng2024agentstudio,zheng2024gpt,wu2024atlas} have demonstrated promising performance in web automation tasks, like product searching~\cite{koh2024visualwebarena}, shopping~\cite{zhou2023webarena}, and navigation~\cite{lu2024gui}. However, most existing web agents heavily rely on the pre-trained capabilities of MLLMs, which are difficult to cover large-scale and diverse web task data~\cite{verma2024adaptagent,bu2025limits}. There remains a significant gap between their prior knowledge and the highly varied and dynamic real-world web pages. To address this gap, prior works usually depend on the design of manually crafted execution pipelines~\cite{yao2023react,koh2024tree,gu2024your} or on the use of human-annotated expert trajectories~\cite{xu2024aguvis,hong2024cogagent,xie2024osworld,chen2024edge} for fine-tuning web agents.

\begin{figure}[t]
\centering
\includegraphics[width=0.9\columnwidth]{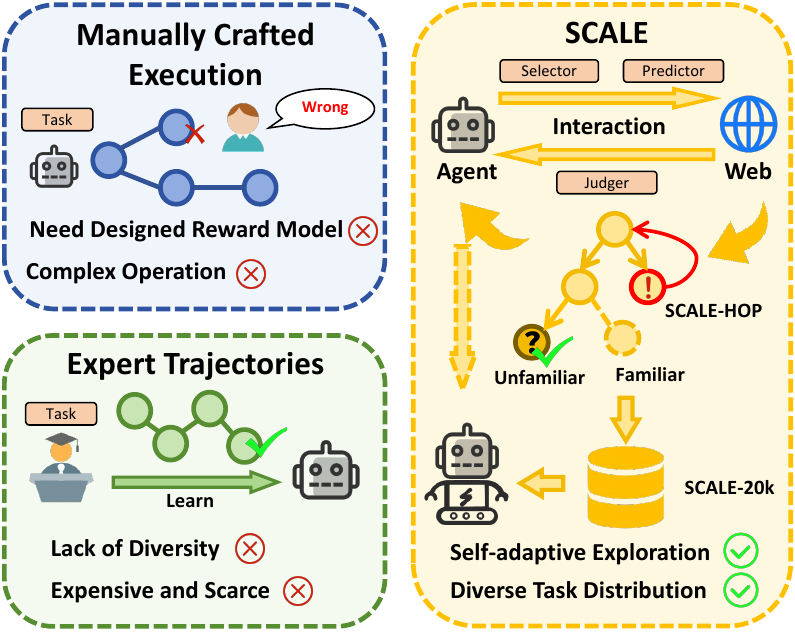} 
\vspace{-5pt}
\caption{\textbf{A comparison between prior methods and our SCALE framework.} SCALE enables autonomous exploration with diverse and scalable task generation, overcoming the limitation in previous approaches.}
\label{fig1}
\vspace{-15pt}
\end{figure}

However, these two types of paradigms mentioned above still have notable limitations: \textbf{1) Reliance on crafted execution pipelines or expert trajectories.} These methods either rely on crafted execution pipelines for task execution or depend on costly expert trajectories to fine-tune MLLMs~\cite{sun2024genesis,wu2024atlas,miaoboosting}. Although these pipelines and trajectories are often tailored for specific scenarios, they still struggle to cope with the unpredictability and diversity of real-world web environments~\cite{verma2024adaptagent,he2024webvoyager,wang2024guisurvey}. Agents in these paradigms tend to act passively and develop dependencies on specific task flows, limiting their ability to explore unfamiliar environments and adapt to dynamic real-world websites. \textbf{2) Challenges in evaluating and expanding cognitive boundaries.} Cognitive boundaries are the limitations of an agent’s abilities, referring to items and operations that the agent struggles to understand or decide on based on prior knowledge~\cite{wang2024agentsurvey}. It is significant for agents to improve their abilities. However, existing methods have neglected the exploration of their own cognitive boundaries. Therefore, how to evaluate and expand agent capabilities in a personalized and dynamic manner remains a critical challenge. \textbf{In summary}, developing general web agents requires enabling agents to proactively adapt to novel environments and continuously expanding their own cognitive boundaries, instead of relying on specific systems or predefined trajectories.

To address these challenges, we propose \textbf{SCALE} (\textit{\textbf{S}elf-\textbf{C}ognitive-\textbf{A}ware \textbf{L}earning and \textbf{E}xploration}), a general framework that removes reliance on expert trajectories and handcrafted pipelines. SCALE enables agents to autonomously discover and expand their cognitive boundaries through self-driven interaction in web environments. Since it operates independently of any specific model architecture, it offers broad generalizability across different models. As shown in Figure~\ref{fig1}, the agent proactively explores without external supervision by identifying its own limitations, triggering exploratory actions in uncertain areas, and adapting based on feedback. To evaluate and refine its understanding, SCALE assigns three roles to the same MLLM: \textit{Selector}, \textit{Predictor}, and \textit{Judger}. The \textit{Selector} proposes challenging actions, the \textit{Predictor} anticipates their outcomes, and the \textit{Judger} verifies prediction accuracy after execution. For example, the \textit{Selector} may choose to click a logo, believing the model is unsure about its effect. The \textit{Predictor} then estimates the outcome based on its current knowledge, expecting nothing to happen. However, once executed, the \textit{Judger} observes a jump to the homepage, revealing a gap in the model’s understanding. This forms a closed-loop process in which the \textit{Selector} and \textit{Predictor} engage in an adversarial interaction, with one aiming to challenge the model and the other aiming to predict correctly, while the \textit{Judger} provides feedback to improve both. The iterative process allows the agent to gradually adapt to complex web environments.


Furthermore, we aim to enhance SCALE by extending its exploration capability from local interactions to global reasoning and coordination. We propose the \textbf{SCALE-Hop} strategy to complement it with global planning and control. SCALE-Hop constructs a graph representation of the agent’s interaction history, where each node represents a specific web state. This graph allows the agent to monitor its exploration coverage and make decisions at a higher level. When exploration becomes difficult or complete at a given node, SCALE-Hop evaluates the model's performance and determines whether to exit and transition to another node that has not been fully explored. This mechanism helps the agent efficiently avoid local traps and supports a more comprehensive understanding of the environment.

Based on the raw exploration traces from the SCALE process in 19 real-world websites, we construct \textbf{SCALE-20k}, a large-scale web task dataset. Compared to existing datasets~\cite{sun2024genesis,xie2024osworld,xu2024aguvis}, SCALE-20k offers not only execution trajectories but also diverse task types, including multi-step tasks, single-step decisions, and page comprehension QA, supporting more comprehensive understanding and reasoning over web content.

Through comprehensive experimentation, our SCALE framework reveals that MLLMs can proactively discover unknown aspects of the environment through self-adversarial processes and achieve self-improvement by continuously discovering and exploring these unknown aspects. Compared to strategies that rely on random exploration or follow task instructions rigidly, SCALE encourages the model to collect and update data specifically in regions it does not yet understand. This leads to more informative and error-revealing data, which better supports the model's improvement. Combined with the global route planning of SCALE-Hop, it enables agents to move beyond shallow, localized behaviors and form a more structured and comprehensive view of complex web environments. Experimentally, SCALE achieves 231.8\% improvement for \texttt{InternVL2.5-8B}~\cite{chen2024expanding} and 176.3\% improvement for \texttt{Qwen2.5-VL-7B}~\cite{bai2025qwen2} on average task success rate. Furthermore, the SCALE-20k dataset achieves high scores in task diversity, reasoning quality, and answer quality. Even when applied directly to other MLLMs, SCALE-20k improves their agent capabilities. These results further demonstrate the effectiveness and generality of the SCALE framework.


Our main contributions are summarized as follows:

\begin{itemize}
    \item We propose \textbf{SCALE}, a self-driven learning and exploration framework that enables web agents to autonomously discover and expand their cognitive boundaries without relying on inference systems or expert trajectories.
    \item We introduce \textbf{SCALE-Hop}, a dynamically adjusted exploration strategy according to the evaluation of cognitive boundaries, which facilitates broader content discovery and promotes the collection of more diverse and robust exploration.
    \item We further provide \textbf{SCALE-20k}, a large-scale dataset derived from the SCALE exploration process, offering diverse task types, high-quality reasoning processes and answers to support future research on web agent understanding and reasoning.
\end{itemize}
\begin{figure*}[t]
\centering
\includegraphics[width=0.9\textwidth]{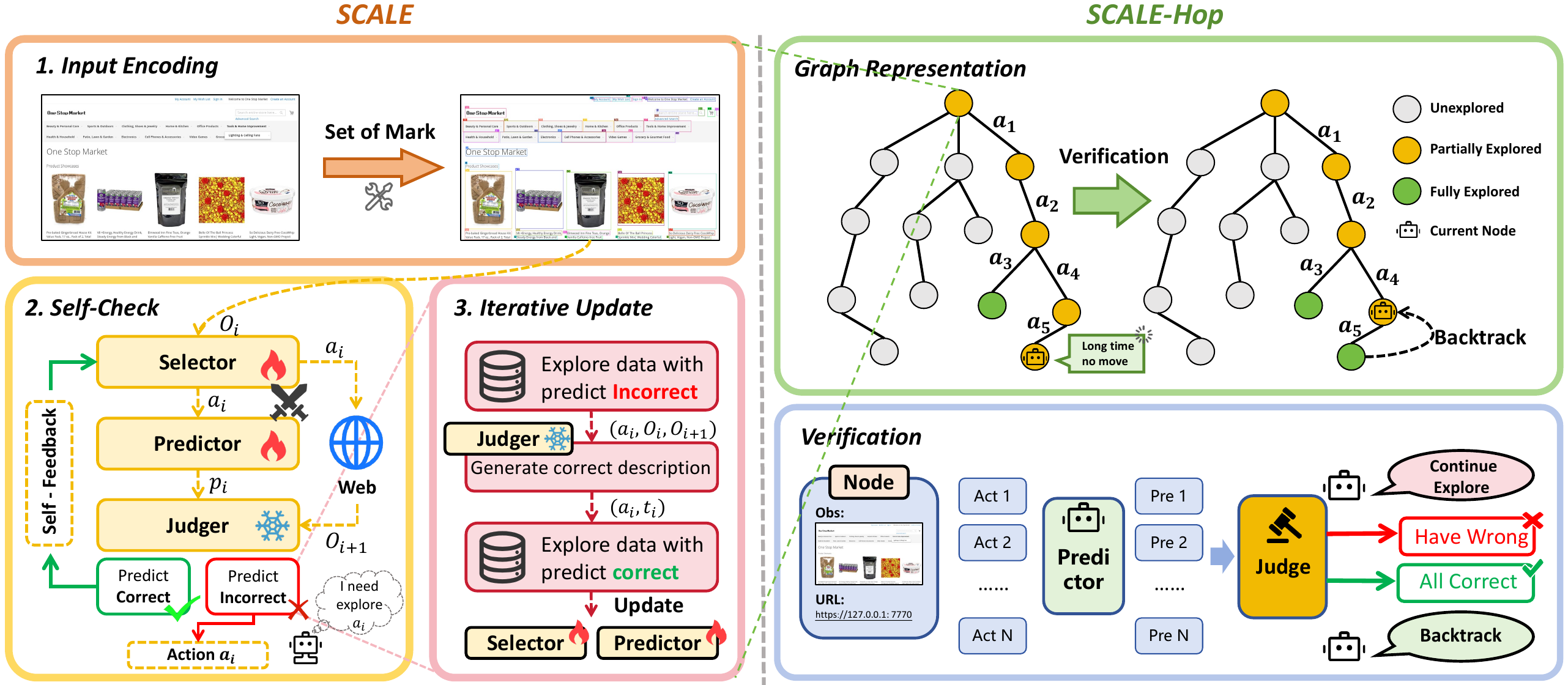} 
\caption{\textbf{The overview of SCALE and SCALE-Hop.} \textbf{SCALE} consists of Input Encoding, Self-Check, and Iterative Update. It enables agents to identify unfamiliar actions, verify predictions, and iteratively improve their reasoning.
\textbf{SCALE-Hop} builds a graph to represent exploration history. It uses verification-guided backtracking to mark nodes as fully explored and guide the agent toward underexplored areas for global navigation.}
\label{fig2}
\vspace{-10pt}
\end{figure*}

\section{Related Work}

\subsection{Crafted Supervision for Web Agents} Early advances in MLLM-based web agents heavily rely on two forms of external supervision: crafted execution pipelines and curated expert trajectories. Crafted pipelines---such as Tree-of-Thought~\cite{koh2024tree}, ReAct~\cite{yao2023react}, and world-model-based planning~\cite{gu2024your}---structure agent reasoning through explicit modules (e.g., search or simulation) to guide actions. While these enhance decision making, they require extensive manual design and lack flexibility beyond pre-defined scenarios.

Meanwhile, approaches using expert trajectories depend on large human-annotated datasets for real-world tasks. Data-centric works like Mind2Web~\cite{deng2023mind2web}, Rico~\cite{deka2017rico}, OSWorld~\cite{xie2024osworld}, and AGUVIS~\cite{xu2024aguvis} advance the field, but face challenges in cost, diversity, and adaptability, resulting in passive agents rigid to new settings. In contrast, our work dispenses with handcrafted modules or costly labels, enabling agents to self-assess and extend their abilities via autonomous exploration.

\subsection{Adaptive Strategy for Web Agents} To address the rigidity of manual supervision, recent methods pursue autonomous adaptation and self-improvement. Frameworks like OpenWebVoyager~\cite{he2024openwebvoyagerbuildingmultimodalweb} and AgentGym~\cite{xi2024agentmgy} use iterative exploration and policy updates~\cite{putta2024agentQ, rafailov2024directpreferenceoptimizationlanguage}, yet usually require pre-defined task templates, limiting adaptation.

Going further, exploration-based strategies such as OS-Genesis~\cite{sun2024genesis} and Learn-by-Interact~\cite{su2025learn} generate trajectories directly through unsupervised interactions, retroactively creating tasks and filtering via reward models. While this reduces task dependence, most methods overlook systematically probing or expanding agent cognitive boundaries. In contrast, our approach focuses on self-cognitive-aware exploration, where agents autonomously identify, diagnose, and broaden their own boundaries for robust environment adaptation.

\section{Method}
In this section, we first introduce the components and process of the SCALE framework, and then further introduce SCALE-Hop's graph representation and verification-guided backtracking mechanism. Finally, we introduce the construction method and analysis of the SCALE-20k dataset.
\subsection{SCALE}
To move beyond handcrafted pipelines and expert trajectories, and to enable agents to proactively identify and expand their own cognitive boundaries in dynamic web environments, we propose a novel framework, \textbf{SCALE}, that enables agents to proactively explore and adapt to web environments without external supervision. In this part, we first introduce its three key components: the \textit{Selector}, \textit{Predictor}, and \textit{Judger}, then describe the three core stages that facilitate cognitive boundary assessment, followed by the formalized execution process of the framework.

\noindent\textbf{Selector, Predictor and Judger.}
The SCALE framework consists of three essential components: \textit{Selector}, \textit{Predictor}, and \textit{Judger}. These roles work together to form a self-confrontation and self-verification mechanism, enabling the agent to continuously identify and focus on actions that are the most challenging to interpret within the current environment. Importantly, all three roles are assumed and executed by the agent itself, forming a fully self-contained reasoning and evaluation loop.

The \textit{Selector} is responsible for identifying elements in the environment that are likely confusing or rarely encountered by the agent, and generating corresponding exploratory actions. For example, in a shopping page, clicking a product is common and easy to understand. In contrast, clicking the site logo is rare and less familiar. The \textit{Selector} prefers the latter to challenge the agent. Then, the \textit{Predictor} is responsible for analyzing the action and anticipating its expected outcome along with an explanation. This enables the system to assess whether the agent knows about the consequences of its decisions before execution. After the action is executed, the \textit{Judger} is responsible for comparing the predicted outcomes to the actual results to determine whether the agent truly understands the selected actions. These three roles interact in a closed-loop that forms the foundation of the SCALE.

\noindent\textbf{The process of SCALE.}
As illustrated in Figure~\ref{fig2}, the operation of SCALE consists of three main stages: Input Encoding, Self-Check, and Iterative Update. These stages work together to support autonomous exploration, cognitive boundary assessment, and continual adaptation.

\textbf{Stage 1: Input Encoding} In this stage, we adopt the commonly used Set-of-Mark (SOM) technique~\cite{yang2023set} to process Graphical User Interface (GUI) images. Unlike text-based methods that rely on accessibility trees or HTML structures, our design processes purely visual information. This choice is motivated by the fact that many real-world websites restrict access to DOM or structural metadata, whereas GUI images are always accessible. This design not only improves the generalization ability of SCALE but also aligns more closely with the way humans perceive and interact with the web through vision.

\textbf{Stage 2: Self-Check} In this stage, the \textit{Selector}, \textit{Predictor}, and \textit{Judger} collaborate to probe the agent’s current understanding of the environment and actively search for its cognitive limits. The \textit{Selector} deliberately proposes challenging actions, targeting uncertain or unfamiliar elements, whose outcomes the agent is least confident about. While the \textit{Predictor} attempts to anticipate the consequence of each proposed action and explain why that outcome should occur based on its existing knowledge. This creates an internal adversarial dynamic: the \textit{Selector} tries to expose behaviors the agent cannot yet understand, while the \textit{Predictor} strives to refute it by accurately forecasting the true outcome. After the action is executed, the \textit{Judger} compares the predicted and observed results to decide whether the agent truly understands the action’s effect. This self-supervised feedback loop reveals where the agent’s understanding breaks down and localizes the cognitive boundaries.

\textbf{Stage 3: Iterative Update} In this stage, the agent refines its reasoning through iterative self-confrontation. Based on the judgments from the \textit{Judger}, actions with mismatched predictions are identified as failure cases, which reveal aspects of the environment that the agent does not understand. We focus on these failure cases rather than successful ones because they expose the agent’s current cognitive limitations and provide the greatest learning signal. These cases are then used to update the \textit{Predictor}, allowing it to learn the actual consequences of previously misunderstood actions and acquire new knowledge about the environment. Meanwhile, these cases also update the \textit{Selector}, guiding it toward actions that remain challenging. Through this mutually reinforcing cycle, the \textit{Selector} and \textit{Predictor} co-evolve in a self-improving manner, continually discovering new cognitive boundaries and expanding the agent’s overall reasoning capability.

\noindent\textbf{Formalized Execution Process.}
We denote the environment at step $i$ as $S_i$, the observation as $O_i$, and the action as $a_i$, where $i \in [1, K]$, and $K$ is the number of interaction steps in a single self-update iteration. After $K$ steps, the agent undergoes one round of self-update. Let $j \in [1, N]$ denote the number of self-update rounds, with $N$ as the maximum number of updates allowed.
For each inner-loop iteration, the \textit{Selector} first receives the current observation $O_i$ and generates an action $a_i$:
\[
a_i, r_{sel_i} = \pi_{sel_j}(O_i)
\]

where $r_{sel_i}$ is the reason for selecting $a_i$. This action is then passed to the \textit{Predictor}, which predicts its outcome:
\[
p_i, r_{pre_i} = \pi_{pre_j}(O_i, a_i, r_{sel_i})
\]

where $p_i$ is the prediction of $a_i$, $r_{pre_i}$is the reason for this prediction.

The action $a_i$ is then executed in the real environment $S_{i+1} = T(S_i, a_i)$ and the agent receives the new observation $O_{i+1} = \Omega(S_{i+1}).$ Next, $O_i$, $O_{i+1}$, and the predicted outcome $p_i$ with explanation $r_{pre_i}$ are passed to the \textit{Judger} to determine whether the prediction was correct:
\[
j_i = \text{Judger}(O_i, O_{i+1}, a_i, p_i, r_{pre_i})
\]

where $j_i \in \{0, 1\}$. If $j_i = 0$, indicating that the prediction was incorrect, the action is considered unfamiliar to the agent and likely lies beyond its current knowledge boundary. In this case, we use the \textit{Judger} to generate the true outcome and explanation:
\[
t_i, r_{t_i} = \text{Judger}(O_i, O_{i+1}, a_i, r_{sel_i})
\]

where $t_i$ is the true description of the result of $a_i$, $r_{t_i}$is the true reason for this description. Then we store this exploratory data as $\text{ExploreData}_j^i = \langle O_i, a_i, r_{sel_i}, t_i, r_{t_i} \rangle$.

If $j_i = 1$, the action is deemed understood and provides no cognitive improvement. The environment is then reset to $S_i$, and the \textit{Selector} generates a new action $a_i'$:
\[
a_i', r_{sel_i}' = \pi_{sel_j}(O_i, a_i, r_{sel_i})
\]

and the above process is repeated until an unfamiliar action ($j_i = 0$) is generated.

After $K$ steps, all collected data $\text{ExploreData}_j = \{\text{ExploreData}_j^i \}_{i=1}^{K}$ is used to fine-tune the \textit{Selector} and \textit{Predictor}:
\[\pi_{sel_{j+1}} = \text{SFT}(\pi_{sel_j}, \text{ExploreData}_j)\]
\[\pi_{pre_{j+1}} = \text{SFT}(\pi_{pre_j}, \text{ExploreData}_j).\]

Since the ability of judging is not trained during this process, the \textit{Judger} can be considered fixed during the iterative updates. The updated policies $\pi_{sel_{j+1}}$ and $\pi_{pre_{j+1}}$ are then used for the next round of exploration, enabling an iterative self-improvement loop.

Through this process, the agent learns from unfamiliar knowledge and expands its cognitive boundaries. Unlike task-driven methods such as ReAct or tree search, SCALE performs self-driven exploration to proactively adapt to new environments and discover boundary cases, enabling higher-quality learning without extra inference-time operations. The full iterative process algorithm and more details can be found in Appendix~A.

\begin{figure*}[t]
\centering
\includegraphics[width=0.9\textwidth]{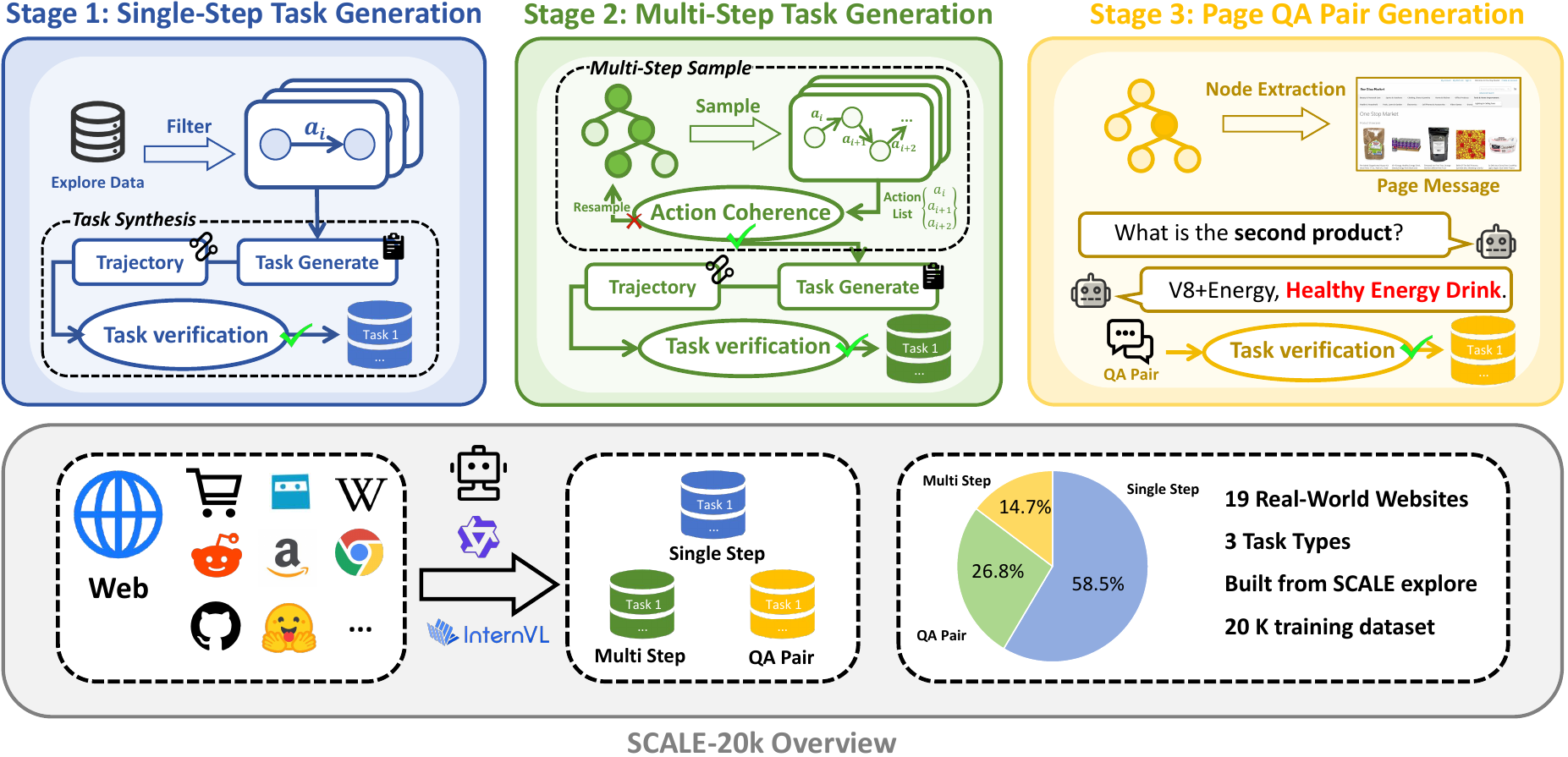} 
\vspace{-7pt}
\caption{\textbf{Overview of the SCALE-20k construction pipeline and composition.} The dataset is constructed in three stages: 1) Single-step tasks are reverse-generated from valid exploration steps; 2) Multi-step tasks are synthesized from coherent trajectories extracted via SCALE-Hop graphs; 3) Page QA pairs are created to test content comprehension. The dataset includes 19 real-world websites and supports three task types, providing rich and diverse supervision for training web agents.}
\label{Overview of the SCALE-20k}

\vspace{-8pt}
\end{figure*}

\subsection{SCALE-Hop}
Exploration in web environments involves not only exploring individual pages and actions but also perceiving and navigating the environment as a whole. However, relying solely on SCALE makes it difficult to obtain a global perspective for exploration and correction. Accordingly, we further propose \textbf{SCALE-Hop}, a global exploration strategy with high-level planning and dynamic control. This section introduces its Graph Representation and Verification-Guided Backtracking mechanism.

\noindent\textbf{Graph Representation.}
As shown in Figure~\ref{fig2}, SCALE-Hop models the agent’s exploration as a directed graph $\mathcal{G} = (\mathcal{N}, \mathcal{E})$, where each node $n_i = (O_i, u_i)$ represents a unique observation $O_i$ and its URL $u_i$, and each edge $e = (n_i, a, n_j)$ denotes a transition from $n_i$ to $n_j$ triggered by an action $a$, where $i, j \in [0, K]$, and $K$ is the total number of nodes. To distinguish different nodes, both the URL and the observation content are considered. For a candidate environment, we first check whether its URL has appeared before. If the URL is new, we directly create a new node $n_{\text{new}} = (O_{\text{new}}, u_{\text{new}})$ and add it to the graph. If the URL already exists, we compare $O_{\text{new}}$ with all observations of nodes sharing the same URL using the Structural Similarity Index (SSIM), only when all SSIM scores are below a predefined threshold $\delta$ do we treat it as a new environment and insert a new node into the graph. During exploration, the graph is dynamically constructed and refined. It provides a compact and expressive way to record the agent’s exploration history, enabling efficient tracking of visited regions, recognition of novel states, and subsequent global planning over the environment.

\noindent\textbf{Verification-Guided Backtracking.}
To further enhance the agent’s global exploration capability, we propose the Verification-Guided Backtracking mechanism as a core component of SCALE-Hop. Each node in the exploration graph is dynamically assigned one of three states: \textit{Unexplored}, \textit{Partially Explored}, or \textit{Fully Explored}, reflecting the agent’s current understanding of that local environment. The transition between these states is driven by continuous verification of prediction accuracy and exploration coverage.

When local exploration stalls or appears complete, a verification phase is triggered. The agent samples $N$ random actions from the current node and uses the \textit{Predictor} to predict their outcomes. If all predictions match the actual results, the node is marked as \textit{Fully Explored}, otherwise, it remains \textit{Partially Explored} for further investigation. Once the verification is passed, the agent then backtracks to the nearest node that is still \textit{Unexplored} or \textit{Partially Explored}. This mechanism balances broad exploration with strategic focus, helping the agent make steady and meaningful progress across diverse web environments.

Through the combination of Graph Representation and Verification-Guided Backtracking, SCALE-Hop performs logically consistent and globally aware exploration, avoiding getting trapped in local regions and missing globally valuable areas. This design enables the SCALE framework to maintain comprehensive and balanced coverage across the entire web environment.

\subsection{SCALE-20k}
Since high-quality web task data are both scarce and expensive to obtain, based on the exploration conducted by SCALE across 19 real-world websites, we construct the \textbf{SCALE-20k} dataset. We first introduce three stages for constructing the dataset, and then analyze its composition and statistics.

\noindent\textbf{Dataset Construction Pipeline.}
As shown in Figure~\ref{Overview of the SCALE-20k}, based on the exploration data obtained from SCALE, we leverage GPT-4o to assist in task construction and validation, and construct the dataset through the following three stages.

\textbf{Stage 1: Single-Step Task Generation} From the SCALE exploration process, we collect numerous valid exploration actions. Then we reverse-construct a corresponding single-step task and reasoning processes based on this action.

\textbf{Stage 2: Multi-Step Task Generation} To reflect the complexity of real-world tasks, we extract coherent multi-step trajectories from the SCALE-Hop graph. The logically validated trajectory is then used to reverse-construct the corresponding task and reasoning process.

\textbf{Stage 3: Page Comprehension QA Pair Generation} To enhance the agent's understanding of web environments, we generate question-answer pairs for each node in the SCALE-Hop graph, providing rich page-level supervision that strengthens the agent’s comprehension and reasoning ability.

\noindent\textbf{Dataset Composition and Statistics.}
As shown in Figure~\ref{Overview of the SCALE-20k}, SCALE-20k comprises exploration data collected by \texttt{Qwen2.5-VL-7B} and \texttt{InternVL2.5-8B} across 19 real-world websites. It includes 15042 single-step tasks, 3780 multi-step tasks, and 6886 page comprehension QA tasks.


\section{Experiments}
In this section, we first describe the experimental setup, then present the main results, followed by an in-depth analysis of the experiments.

\subsection{Experimental Setup}
\noindent\textbf{Baselines.} We select \texttt{Qwen2.5-VL-7B}~\cite{bai2025qwen2} and \texttt{InternVL2.5-8B}~\cite{chen2024expanding} as backbone models to evaluate our method. We used GPT-4o as a representative closed-source baseline. We also compare against strong open-source agents, including Augvis, a vision-only GUI agent~\cite{xu2024aguvis}, and ViGoRL, a vision-based reinforcement learning agent~\cite{sarch2025grounded}. Beyond zero-shot settings, we consider three enhanced strategies: \textbf{1)} Imitation learning on GPT-4o successful trajectories, \textbf{2)} OS-Genesis exploration method~\cite{sun2024genesis}, which constructs tasks from randomly explored trajectories, and \textbf{3)} Tree Search strategy that explores multiple reasoning paths during inference time via a tree structured search. To further validate the effectiveness of our proposed SCALE-20k dataset, we additionally include a model that is entirely unrelated to our exploration model, \texttt{LLaVA‑NeXT‑8B}~\cite{li2024llava}, and fine-tune it directly on SCALE-20k.

\noindent\textbf{Evaluation Benchmarks.}
We conduct evaluations on two widely-used web agent benchmarks: \textbf{VisualWebArena} and \textbf{WebVoyager}. VisualWebArena covers diverse domains such as shopping, classifieds, and forums, while WebVoyager includes real-world websites with complex layouts and dynamic, real-time content changes. For WebVoyager, we test on the first 11 tasks of each website to ensure consistency. To enable a fair comparison across all methods, we unify the input format by using only webpage screenshots with visual markup as the observation. All other auxiliary information, such as accessibility trees and page descriptions, is removed. Further experimental details are provided in Appendix B.

\begin{table*}[ht]
\centering
\tiny
\renewcommand{\arraystretch}{1.1}
\resizebox{0.95\textwidth}{!}{
\begin{tabular}{cccccccccc}
\toprule
\multirow{2}{*}{Model} & \multirow{2}{*}{Strategy} & \multicolumn{2}{c}{Shopping} & \multicolumn{2}{c}{Classifieds} & \multicolumn{2}{c}{Reddit} & \multicolumn{2}{c}{WebVoyager} \\
 &  & SR$\uparrow$ & AS$\downarrow$ & SR$\uparrow$ & AS$\downarrow$ & SR$\uparrow$ & AS$\downarrow$ & SR$\uparrow$ & AS$\downarrow$ \\
\midrule
GPT-4o     & Zero-Shot           & \textbf{17.2} & 15.5 & \textbf{13.7} & 12.5 & \textbf{6.7} & 17.6 & \textbf{9.6} & \textbf{14.7} \\
\midrule
Augvis     & Zero-Shot           & 4.3  & - & 0.4 & - & 0.5 & - & 0.0 & - \\
ViGoRL     & Zero-Shot           & 5.6  & 13.4 & 6.9 & 12.4 & 2.9 & 14.6 & 0.6 & 14.9 \\
\midrule
\multirow{5}{*}{InternVL2.5-8B}
 & Zero-Shot           & 3.9  & 23.5 & 0.4 & 22.7 & 1.4 & 22.7 & 0.0 & - \\
 & GPT Trajectory Imitation & 8.8  & 16.5 & 3.4 & 16.7 & 3.0 & 17.2 & -   & - \\
 & OS-Genesis     & \textbf{11.6} & \underline{6.6} & 3.0 & \underline{6.3} & \textbf{4.3} & \underline{9.7} & 0.0 & - \\
 & Tree Search         & 3.0  & - & \underline{4.3} & - & 1.0 & - & -   & - \\
 & \textbf{SCALE(ours)}                & \underline{11.0} & \textbf{4.9} & \textbf{6.4} & \textbf{5.5} & \underline{3.3} & \textbf{7.2} & \textbf{1.8} & \textbf{14.9} \\
\midrule
\multirow{5}{*}{Qwen2.5-VL-7B}
 & Zero-Shot           & 4.1  & 19.4 & 6.0  & 15.8 & 2.4 & 16.1 & 0.6 & \textbf{15.0} \\
 & GPT Trajectory Imitation & \textbf{18.3} & \underline{8.5} & \underline{10.7} & \underline{9.2} & \underline{3.3} & \underline{12.0} & -   & - \\
 & OS-Genesis    & 11.2 & 12.7 & 8.6  & 11.3 & 1.4 & 12.7 & \underline{6.7} & \textbf{15.0} \\
 & Tree Search         & 4.1  & - & 1.7  & - & 1.0 & - & -   & - \\
 & \textbf{SCALE(ours)}                & \underline{14.4} & \textbf{5.7} & \textbf{12.0} & \textbf{4.1} & \textbf{4.8} & \textbf{8.4} & \textbf{7.9} & \textbf{15.0} \\
\midrule
\multirow{2}{*}{LLaVA‑NeXT‑8B} 
    & Zero-Shot      & 0 & - & 0 & - & 1.4 & \textbf{4.6} & 0.0 & - \\
    & \textbf{SCALE-20k}      & \textbf{2.1} & \textbf{10.1} & \textbf{0.8} & \textbf{8.2} & \textbf{1.9} & 9.6 & 0.0 & - \\

\bottomrule
\end{tabular}
}
\caption{\textbf{Overall Performance Comparison Across Models and Exploration Strategies.} Success rate (SR, \%) and average steps (AS) of different strategies across models on multiple web benchmarks. SR is reported as a percentage without the \% symbol. The best value in each group is shown in \textbf{bold}, and the second-best value is \underline{underlined}. SCALE denotes each model’s adapted performance via our SCALE procedure, while SCALE-20k refers to results obtained by training directly on the constructed SCALE-20k dataset.}
\label{tab:main_results}

\vspace{-5pt}
\end{table*}


\begin{table}[t]
\centering
\renewcommand{\arraystretch}{1.2}
\resizebox{\columnwidth}{!}{  
\begin{tabular}{llcccc}
\hline
\toprule
Model          & Strategy         & Shopping & Classifieds & Reddit & Overall\\ \hline
\multirow{4}{*}{Qwen2.5-VL-7B}
               & Zero-Shot        &  4.5              &    6.0          &  1.9   & 4.1\\
               & SCALE (20-25)    &  6.7              &    \underline{9.8}          &  \textbf{5.3}  & 7.2 \\
               & SCALE (40-25)    &  \underline{9.0}              &    9.0          &   4.3  & \underline{7.9}\\
               & SCALE (60-25)    &  \textbf{14.4}    &    \textbf{12.0}         &   \underline{4.8}  & \textbf{11.9}\\ \hline
\end{tabular}
}
\caption{\textbf{Performance of Qwen2.5-VL-7B under different exploration levels.} SR is shown as a percentage without the \% symbol. SCALE (X-25) denotes X outer-loop hops and 25 inner-loop steps. Best results are in \textbf{bold}, second-best are \underline{underlined}.}
\label{tab:scale_depth}
\end{table}

\begin{table}[t]
\centering
\small
\renewcommand{\arraystretch}{1.1}
\resizebox{\columnwidth}{!}{ 
\begin{tabular}{ccccccccc}
\hline
\toprule
\multirow{2}{*}{Strategy} & \multicolumn{2}{c}{Shopping} & \multicolumn{2}{c}{Classifieds} & \multicolumn{2}{c}{Reddit} & \multicolumn{2}{c}{Overall}\\ 
                          & SR             & Node          & SR            & Node            & SR          & Node         & SR          & Node \\ \hline
Random Walk               & \textbf{14.8}          & 207           & \underline{7.3}     & \underline{163}             & 3.8      & \underline{29}  & \underline{10.4}   & \underline{399}        \\ \hline
w/o SCALE-Hop             & 13.5         & \underline{226}           & 6.0       & 25              & \textbf{7.1}    & 26       & 10.1   & 277  \\ \hline
SCALE                     & \underline{14.4}        & \textbf{347}  & \textbf{12.0}       & \textbf{292}    & \underline{4.8}      & \textbf{237}  & \textbf{11.6}   & \textbf{876}\\ \hline
\end{tabular}
}
\caption{\textbf{Ablation study of exploration strategies across three web domains.} We report task success rate (SR) and number of visited nodes (Node) to reflect effectiveness and coverage. SR values omit the \% symbol. Best results are in \textbf{bold}, second-best are \underline{underlined}.}
\label{tab:ablation_scalehop}
\vspace{-15pt}
\end{table}

\subsection{Main Results}
Through comprehensive experimentation, we validate the effectiveness and generalization of our proposed method:

\noindent\textbf{Effectiveness of SCALE.}
The SCALE framework is effective across MLLMs of different scales. As shown in Table~\ref{tab:main_results}, both \texttt{InternVL2.5-8B} and \texttt{Qwen2.5-VL-7B} demonstrate the ability to autonomously explore web environments in SCALE. As a result, they achieve average performance improvements of 231.8\% and 176.3\% over their vanilla counterparts. In addition, SCALE significantly outperforms other open-source vision-language agents such as Augvis, and ViGoRL across most environments. This validates the effectiveness of SCALE in enhancing the agent's capabilities. Moreover, the consistent improvements across different models suggest that our method is general and can be applied to a wide range of MLLM backbones. 

Beyond vanilla zero-shot baselines, SCALE outperforms two mainstream improvement paradigms—expert-trajectory supervision and manually engineered reasoning pipelines. Compared with models trained on GPT‑4o trajectories or OS‑Genesis exploration data, SCALE achieves consistently better or competitive performance, often reaching or surpassing state-of-the-art results. Its advantage is particularly notable on WebVoyager, where dynamic content and diverse page layouts make static or manually curated trajectories insufficient, highlighting SCALE’s stronger generalization and adaptability.

In terms of efficiency, we report the average number of steps (AS) taken across all tasks, regardless of success. As shown in Table~\ref{tab:main_results}, SCALE achieves the lowest or second-lowest AS in most domains, indicating more concise and effective reasoning paths compared to trajectory-based or tree-search baselines.

Moreover, as shown in Table~\ref{tab:scale_depth}, increasing the number of SCALE iterations leads to consistent performance gains, indicating that deeper exploration enables the agent to discover more informative behaviors and improve task success rates.

\begin{figure*}[t]
\centering
\includegraphics[width=0.9\textwidth]{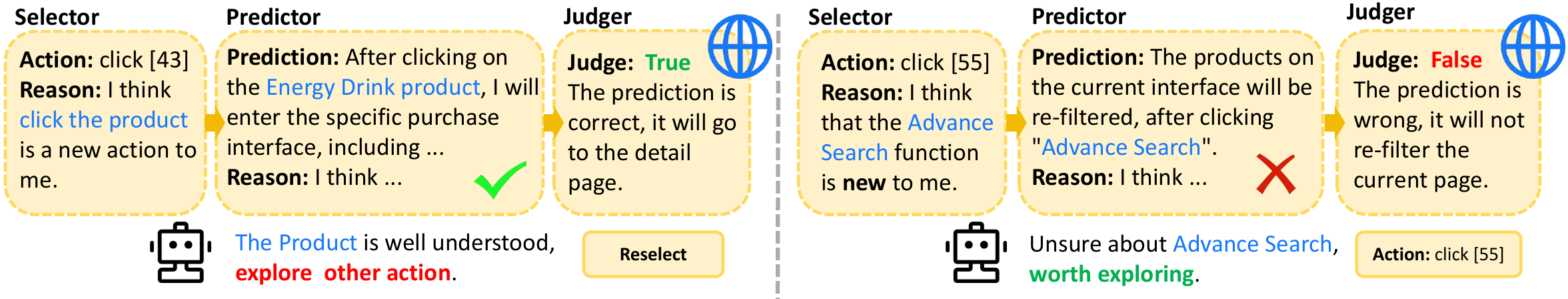} 
\vspace{-8pt}
\caption{A case of cognitive boundary discovery by the \textit{Selector}, \textit{Predictor} and \textit{Judger} in SCALE}
\label{fig4}
\vspace{-8pt}
\end{figure*}

\begin{figure}[t]
\centering
\includegraphics[width=0.9\columnwidth]{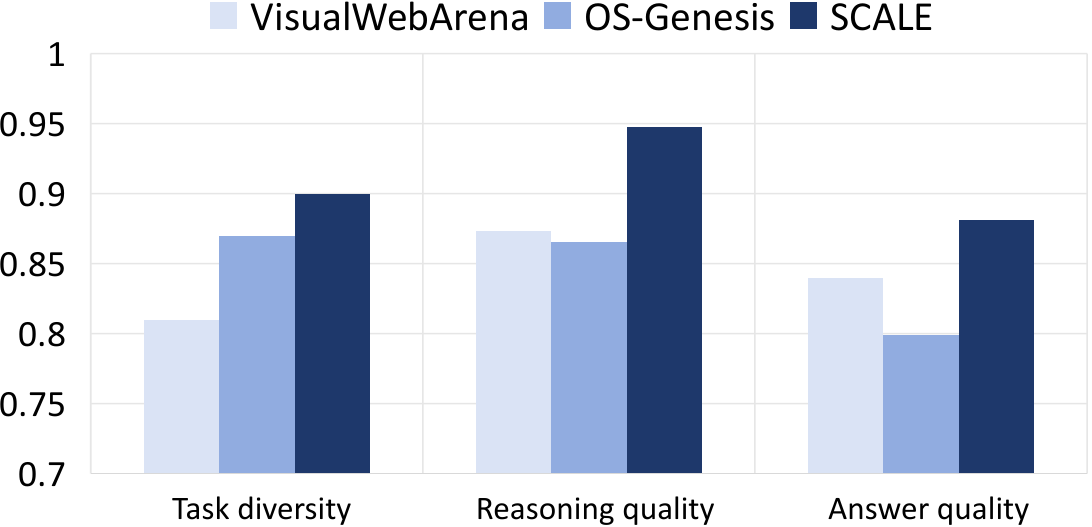} 
\vspace{-5pt}
\caption{Comparison of SCALE-20k, OS-Genesis, and VisualWebArena across three dimensions.}
\label{diversity Comparison}
\vspace{-10pt}
\end{figure}

\begin{figure}[t]
\centering
\begin{subfigure}[t]{0.48\columnwidth}
    \centering
    \includegraphics[width=\linewidth]{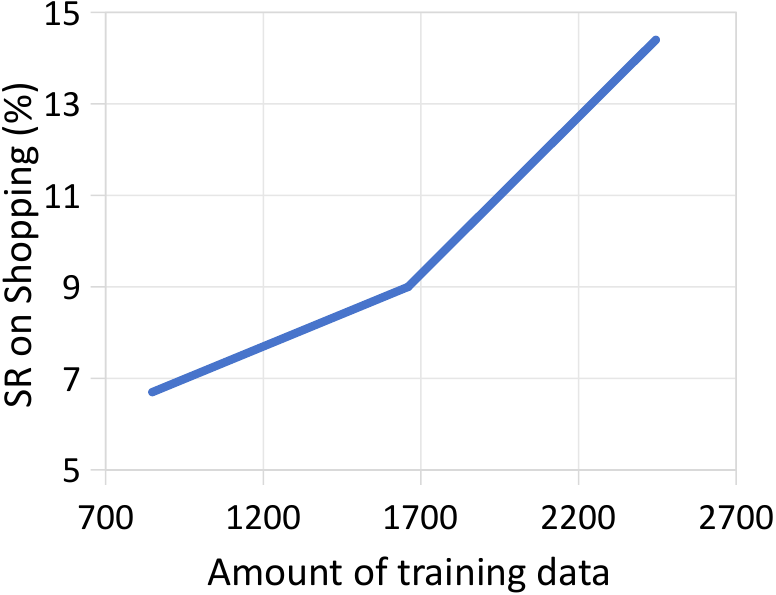}
    \caption{ }
    \label{fig6a}
\end{subfigure}
\hfill
\begin{subfigure}[t]{0.48\columnwidth}
    \centering
    \includegraphics[width=\linewidth]{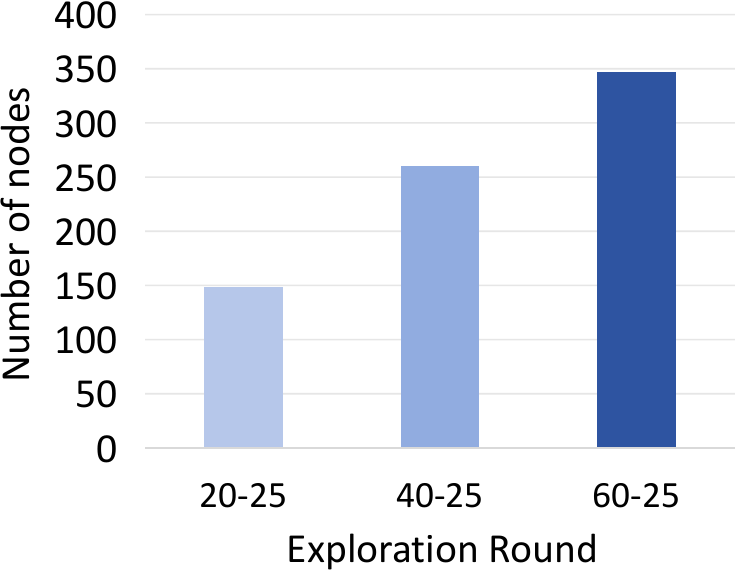}
    \caption{ }
    \label{fig6b}
\end{subfigure}
\vspace{-8pt}
\caption{(a) Impact of training data size on Success Rate in shopping environment. (b) Node diversity under different exploration rounds. (X-25 indicates X outer-loops and 25 inner steps.)}
\label{fig6}
\vspace{-15pt}
\end{figure}

\subsection{In-Depth Analysis}
In this section, we further analyze how SCALE enables deeper and broader exploration.

\noindent\textbf{Ablation Study: The Importance of Self-Adversarial Interaction in Exploration.}  
We observe that the self-adversarial mechanism in SCALE plays a vital role in enhancing exploration behavior. As shown in Table~\ref{tab:ablation_scalehop}, our method significantly outperforms random walk in both task success rate (SR) and exploration coverage. It consistently visits more unique nodes with higher diversity in URL distribution and visual layout. Many of these nodes lie in rarely accessed regions—such as footer menus, hidden tabs, and domain edges—that are often overlooked by goal-driven or randomly sampling agents. This demonstrates that self-adversarial interaction is key to enabling not only deeper exploration, but also continual self-driven improvement.

\noindent\textbf{Ablation Study: Contribution of SCALE-Hop.}  
Table~\ref{tab:ablation_scalehop} further highlights the impact of SCALE-Hop within the framework. Its integration yields the largest performance improvement across environments, boosting both SR and the number of visited nodes. On average, SCALE-Hop increases node coverage by 216\% compared to the w/o SCALE-Hop setting, enabling the agent to escape local patterns and uncover a wider range of states. This confirms SCALE-Hop as a crucial driver of long-horizon, diverse, efficient,  and self-improving exploration.

\noindent\textbf{Diversity of the SCALE-20k.} 
To evaluate dataset diversity, we compute the average cosine distance between Sentence‑BERT~\cite{reimers-gurevych-2019-sentence} embeddings of task descriptions. In addition, we assess dataset quality using GPT‑4o along two complementary dimensions, \textit{reasoning quality} and \textit{answer quality}, with evaluation prompts detailed in Appendix C. As shown in Figure~\ref{diversity Comparison}, SCALE‑20k achieves strong performance across all three metrics, exhibiting higher semantic diversity and better reasoning-action correctness than other datasets. Moreover, as shown in Table~\ref{tab:main_results}, even a model unrelated to our framework (LLaVA‑NeXT‑8B) gains clear improvements when fine‑tuned solely on SCALE‑20k, further demonstrating the effectiveness and generality of our dataset.

\noindent\textbf{Case Study.} As shown in Figure~\ref{fig4}, this case demonstrates how SCALE's components collaboratively identify the agent's cognitive boundary. On the left, although the selected action (clicking the product) is valid, the Predictor already understands its outcome well, suggesting this interaction is within the agent's current knowledge and offers limited exploratory value. On the right, a new action (clicking “Advance Search”) is selected. The Predictor's expectation is invalidated by the Judger, signaling a gap in understanding. This mismatch reveals a gap in understanding and drives targeted exploration, showcasing SCALE’s ability to recognize and expand cognitive boundaries.



\noindent\textbf{Towards Lifelong Exploration via Self-Adaptation.}
As exploration progresses, the agent gradually improves its ability to handle complex interfaces without external supervision—each update relies solely on internal judgment and feedback. As shown in Figure~\ref{fig6}, both task success rate and exploration diversity increase steadily. This illustrates a promising direction toward lifelong exploration learning, where agents continuously adapt to evolving web environments through self-driven feedback loops.
\section{Conclusion}
In this work, we propose \textbf{SCALE}, a self-driven framework that enables web agents to autonomously explore and expand their cognitive boundaries without relying on handcrafted pipelines or expert demonstrations. It uses the \textit{Selector}, \textit{Predictor}, and \textit{Judger} to collaboratively form internal feedback loops, enabling the agent to identify and overcome its cognitive limitations. At the global level, we design the \textbf{SCALE-Hop} strategy to guide broader and more strategic exploration across diverse web environments. Additionally, we construct \textbf{SCALE-20k}, a large-scale and diverse dataset, from SCALE trajectories. Experimental results demonstrate the effectiveness and adaptability of our approach for building general web agents.
\section*{Acknowledgment}
This work was supported by the National Key Research and Development Program of China (2025ZD0123100), the Zhejiang NSF (LQK26F020001), Key R\&D Program of Zhejiang (2026SDXT005), Fundamental Research Funds for the Central Universities (226-2025-00057), Ningbo Yongjiang Talent Introduction Programme (2024A-401-G), Zhejiang University Education Foundation Qizhen Scholar Foundation.


{
    \small
    \bibliographystyle{ieeenat_fullname}
    \bibliography{main}

@String(ICLR = {Int. Conf. Learn. Represent.})

@String(ICLR  = {ICLR})

@article{xu2024aguvis,
  title={Aguvis: Unified Pure Vision Agents for Autonomous GUI Interaction},
  author={Xu, Yiheng and Wang, Zekun and Wang, Junli and Lu, Dunjie and Xie, Tianbao and Saha, Amrita and Sahoo, Doyen and Yu, Tao and Xiong, Caiming},
  journal={arXiv preprint arXiv:2412.04454},
  year={2024}
}

@inproceedings{lin2025showui,
  title={Showui: One vision-language-action model for gui visual agent},
  author={Lin, Kevin Qinghong and Li, Linjie and Gao, Difei and Yang, Zhengyuan and Wu, Shiwei and Bai, Zechen and Lei, Stan Weixian and Wang, Lijuan and Shou, Mike Zheng},
  booktitle={Proceedings of the Computer Vision and Pattern Recognition Conference},
  pages={19498--19508},
  year={2025}
}

@article{lu2024gui,
  title={Gui odyssey: A comprehensive dataset for cross-app gui navigation on mobile devices},
  author={Lu, Quanfeng and Shao, Wenqi and Liu, Zitao and Meng, Fanqing and Li, Boxuan and Chen, Botong and Huang, Siyuan and Zhang, Kaipeng and Qiao, Yu and Luo, Ping},
  journal={arXiv preprint arXiv:2406.08451},
  year={2024}
}

@article{zhou2023webarena,
  title={Webarena: A realistic web environment for building autonomous agents},
  author={Zhou, Shuyan and Xu, Frank F and Zhu, Hao and Zhou, Xuhui and Lo, Robert and Sridhar, Abishek and Cheng, Xianyi and Ou, Tianyue and Bisk, Yonatan and Fried, Daniel and others},
  journal={arXiv preprint arXiv:2307.13854},
  year={2023}
}

@article{koh2024visualwebarena,
  title={Visualwebarena: Evaluating multimodal agents on realistic visual web tasks},
  author={Koh, Jing Yu and Lo, Robert and Jang, Lawrence and Duvvur, Vikram and Lim, Ming Chong and Huang, Po-Yu and Neubig, Graham and Zhou, Shuyan and Salakhutdinov, Ruslan and Fried, Daniel},
  journal={arXiv preprint arXiv:2401.13649},
  year={2024}
}

@article{zheng2024agentstudio,
  title={Agentstudio: A toolkit for building general virtual agents},
  author={Zheng, Longtao and Huang, Zhiyuan and Xue, Zhenghai and Wang, Xinrun and An, Bo and Yan, Shuicheng},
  journal={arXiv preprint arXiv:2403.17918},
  year={2024}
}

@article{verma2024adaptagent,
  title={Adaptagent: Adapting multimodal web agents with few-shot learning from human demonstrations},
  author={Verma, Gaurav and Kaur, Rachneet and Srishankar, Nishan and Zeng, Zhen and Balch, Tucker and Veloso, Manuela},
  journal={arXiv preprint arXiv:2411.13451},
  year={2024}
}

@article{xie2024osworld,
  title={Osworld: Benchmarking multimodal agents for open-ended tasks in real computer environments},
  author={Xie, Tianbao and Zhang, Danyang and Chen, Jixuan and Li, Xiaochuan and Zhao, Siheng and Cao, Ruisheng and Hua, Toh J and Cheng, Zhoujun and Shin, Dongchan and Lei, Fangyu and others},
  journal={Advances in Neural Information Processing Systems},
  volume={37},
  pages={52040--52094},
  year={2024}
}

@inproceedings{hong2024cogagent,
  title={Cogagent: A visual language model for gui agents},
  author={Hong, Wenyi and Wang, Weihan and Lv, Qingsong and Xu, Jiazheng and Yu, Wenmeng and Ji, Junhui and Wang, Yan and Wang, Zihan and Dong, Yuxiao and Ding, Ming and others},
  booktitle={Proceedings of the IEEE/CVF Conference on Computer Vision and Pattern Recognition},
  pages={14281--14290},
  year={2024}
}

@article{zheng2024gpt,
  title={Gpt-4v (ision) is a generalist web agent, if grounded},
  author={Zheng, Boyuan and Gou, Boyu and Kil, Jihyung and Sun, Huan and Su, Yu},
  journal={arXiv preprint arXiv:2401.01614},
  year={2024}
}

@inproceedings{yao2023react,
  title={React: Synergizing reasoning and acting in language models},
  author={Yao, Shunyu and Zhao, Jeffrey and Yu, Dian and Du, Nan and Shafran, Izhak and Narasimhan, Karthik and Cao, Yuan},
  booktitle={International Conference on Learning Representations (ICLR)},
  year={2023}
}

@article{koh2024tree,
  title={Tree search for language model agents},
  author={Koh, Jing Yu and McAleer, Stephen and Fried, Daniel and Salakhutdinov, Ruslan},
  journal={arXiv preprint arXiv:2407.01476},
  year={2024}
}

@article{gu2024your,
  title={Is your llm secretly a world model of the internet? model-based planning for web agents},
  author={Gu, Yu and Zhang, Kai and Ning, Yuting and Zheng, Boyuan and Gou, Boyu and Xue, Tianci and Chang, Cheng and Srivastava, Sanjari and Xie, Yanan and Qi, Peng and others},
  journal={arXiv preprint arXiv:2411.06559},
  year={2024}
}

@article{sun2024genesis,
  title={OS-Genesis: Automating GUI Agent Trajectory Construction via Reverse Task Synthesis},
  author={Sun, Qiushi and Cheng, Kanzhi and Ding, Zichen and Jin, Chuanyang and Wang, Yian and Xu, Fangzhi and Wu, Zhenyu and Jia, Chengyou and Chen, Liheng and Liu, Zhoumianze and others},
  journal={arXiv preprint arXiv:2412.19723},
  year={2024}
}

@article{wu2024atlas,
  title={Os-atlas: A foundation action model for generalist gui agents},
  author={Wu, Zhiyong and Wu, Zhenyu and Xu, Fangzhi and Wang, Yian and Sun, Qiushi and Jia, Chengyou and Cheng, Kanzhi and Ding, Zichen and Chen, Liheng and Liang, Paul Pu and others},
  journal={arXiv preprint arXiv:2410.23218},
  year={2024}
}

@article{he2024webvoyager,
  title={WebVoyager: Building an end-to-end web agent with large multimodal models},
  author={He, Hongliang and Yao, Wenlin and Ma, Kaixin and Yu, Wenhao and Dai, Yong and Zhang, Hongming and Lan, Zhenzhong and Yu, Dong},
  journal={arXiv preprint arXiv:2401.13919},
  year={2024}
}

@article{su2025learn,
  title={Learn-by-interact: A Data-Centric Framework for Self-Adaptive Agents in Realistic Environments},
  author={Su, Hongjin and Sun, Ruoxi and Yoon, Jinsung and Yin, Pengcheng and Yu, Tao and Ar{\i}k, Sercan {\"O}},
  journal={arXiv preprint arXiv:2501.10893},
  year={2025}
}

@article{chen2024edge,
  title={Edge: Enhanced grounded gui understanding with enriched multi-granularity synthetic data},
  author={Chen, Xuetian and Li, Hangcheng and Liang, Jiaqing and Jiang, Sihang and Yang, Deqing},
  journal={arXiv preprint arXiv:2410.19461},
  year={2024}
}

@article{wang2024guisurvey,
  title={Gui agents with foundation models: A comprehensive survey},
  author={Wang, Shuai and Liu, Weiwen and Chen, Jingxuan and Zhou, Yuqi and Gan, Weinan and Zeng, Xingshan and Che, Yuhan and Yu, Shuai and Hao, Xinlong and Shao, Kun and others},
  journal={arXiv preprint arXiv:2411.04890},
  year={2024}
}

@article{putta2024agentQ,
  title={Agent q: Advanced reasoning and learning for autonomous ai agents},
  author={Putta, Pranav and Mills, Edmund and Garg, Naman and Motwani, Sumeet and Finn, Chelsea and Garg, Divyansh and Rafailov, Rafael},
  journal={arXiv preprint arXiv:2408.07199},
  year={2024}
}

@article{wang2024agentsurvey,
  title={A survey on large language model based autonomous agents},
  author={Wang, Lei and Ma, Chen and Feng, Xueyang and Zhang, Zeyu and Yang, Hao and Zhang, Jingsen and Chen, Zhiyuan and Tang, Jiakai and Chen, Xu and Lin, Yankai and others},
  journal={Frontiers of Computer Science},
  volume={18},
  number={6},
  pages={186345},
  year={2024},
  publisher={Springer}
}

@misc{deng2023mind2web,
  title={Mind2Web: Towards a Generalist Agent for the Web},
  author={Xiang Deng and Yu Gu and Boyuan Zheng and Shijie Chen and Samuel Stevens and Boshi Wang and Huan Sun and Yu Su},
  year={2023},
  eprint={2306.06070},
  archivePrefix={arXiv},
  primaryClass={cs.CL}
}

@inproceedings{deka2017rico,
  title={Rico: A mobile app dataset for building data-driven design applications},
  author={Deka, Biplab and Huang, Zifeng and Franzen, Chad and Hibschman, Joshua and Afergan, Daniel and Li, Yang and Nichols, Jeffrey and Kumar, Ranjitha},
  booktitle={Proceedings of the 30th annual ACM symposium on user interface software and technology},
  pages={845--854},
  year={2017}
}

@misc{he2024openwebvoyagerbuildingmultimodalweb,
      title={OpenWebVoyager: Building Multimodal Web Agents via Iterative Real-World Exploration, Feedback and Optimization}, 
      author={Hongliang He and Wenlin Yao and Kaixin Ma and Wenhao Yu and Hongming Zhang and Tianqing Fang and Zhenzhong Lan and Dong Yu},
      year={2024},
      eprint={2410.19609},
      archivePrefix={arXiv},
      primaryClass={cs.CL},
      url={https://arxiv.org/abs/2410.19609}, 
}

@article{xi2024agentmgy,
      title={AgentGym: Evolving Large Language Model-based Agents across Diverse Environments}, 
      author={Zhiheng Xi and Yiwen Ding and Wenxiang Chen and Boyang Hong and Honglin Guo and Junzhe Wang and Dingwen Yang and Chenyang Liao and Xin Guo and Wei He and Songyang Gao and Lu Chen and Rui Zheng and Yicheng Zou and Tao Gui and Qi Zhang and Xipeng Qiu and Xuanjing Huang and Zuxuan Wu and Yu-Gang Jiang},
      year={2024},
      journal={arXiv preprint arXiv:2406.04151}
}

@article{rafailov2024directpreferenceoptimizationlanguage,
  title={Direct preference optimization: Your language model is secretly a reward model},
  author={Rafailov, Rafael and Sharma, Archit and Mitchell, Eric and Manning, Christopher D and Ermon, Stefano and Finn, Chelsea},
  journal={Advances in neural information processing systems},
  volume={36},
  pages={53728--53741},
  year={2023}
}

@article{yang2023set,
  title={Set-of-mark prompting unleashes extraordinary visual grounding in gpt-4v},
  author={Yang, Jianwei and Zhang, Hao and Li, Feng and Zou, Xueyan and Li, Chunyuan and Gao, Jianfeng},
  journal={arXiv preprint arXiv:2310.11441},
  year={2023}
}

@article{sarch2025grounded,
  title={Grounded Reinforcement Learning for Visual Reasoning},
  author={Sarch, Gabriel and Saha, Snigdha and Khandelwal, Naitik and Jain, Ayush and Tarr, Michael J and Kumar, Aviral and Fragkiadaki, Katerina},
  journal={arXiv preprint arXiv:2505.23678},
  year={2025}
}

@inproceedings{reimers-gurevych-2019-sentence,
    title = "Sentence-{BERT}: Sentence Embeddings using {S}iamese {BERT}-Networks",
    author = "Reimers, Nils  and
      Gurevych, Iryna",
    editor = "Inui, Kentaro  and
      Jiang, Jing  and
      Ng, Vincent  and
      Wan, Xiaojun",
    booktitle = "Proceedings of the 2019 Conference on Empirical Methods in Natural Language Processing and the 9th International Joint Conference on Natural Language Processing (EMNLP-IJCNLP)",
    month = nov,
    year = "2019",
    address = "Hong Kong, China",
    publisher = "Association for Computational Linguistics",
    url = "https://aclanthology.org/D19-1410/",
    doi = "10.18653/v1/D19-1410",
    pages = "3982--3992"
}

@article{bai2025qwen2,
  title={Qwen2. 5-vl technical report},
  author={Bai, Shuai and Chen, Keqin and Liu, Xuejing and Wang, Jialin and Ge, Wenbin and Song, Sibo and Dang, Kai and Wang, Peng and Wang, Shijie and Tang, Jun and others},
  journal={arXiv preprint arXiv:2502.13923},
  year={2025}
}

@article{chen2024expanding,
  title={Expanding performance boundaries of open-source multimodal models with model, data, and test-time scaling},
  author={Chen, Zhe and Wang, Weiyun and Cao, Yue and Liu, Yangzhou and Gao, Zhangwei and Cui, Erfei and Zhu, Jinguo and Ye, Shenglong and Tian, Hao and Liu, Zhaoyang and others},
  journal={arXiv preprint arXiv:2412.05271},
  year={2024}
}

@inproceedings{wan2024omniparser,
  title={Omniparser: A unified framework for text spotting key information extraction and table recognition},
  author={Wan, Jianqiang and Song, Sibo and Yu, Wenwen and Liu, Yuliang and Cheng, Wenqing and Huang, Fei and Bai, Xiang and Yao, Cong and Yang, Zhibo},
  booktitle={Proceedings of the IEEE/CVF conference on computer vision and pattern recognition},
  pages={15641--15653},
  year={2024}
}

@article{li2024llava,
  title={Llava-next-interleave: Tackling multi-image, video, and 3d in large multimodal models},
  author={Li, Feng and Zhang, Renrui and Zhang, Hao and Zhang, Yuanhan and Li, Bo and Li, Wei and Ma, Zejun and Li, Chunyuan},
  journal={arXiv preprint arXiv:2407.07895},
  year={2024}
}

@inproceedings{miaoboosting,
  title={Boosting Virtual Agent Learning and Reasoning: A Step-Wise, Multi-Dimensional, and Generalist Reward Model with Benchmark},
  author={Miao, Bingchen and Wu, Yang and Gao, Minghe and Yu, Qifan and Bu, Wendong and Zhang, Wenqiao and Li, Yunfei and Tang, Siliang and Chua, Tat-Seng and Li, Juncheng},
  booktitle={Forty-second International Conference on Machine Learning},
  year={2025}
}

@inproceedings{bu2025limits,
  title={What Limits Virtual Agent Application? OmniBench: A Scalable Multi-Dimensional Benchmark for Essential Virtual Agent Capabilities},
  author={Bu, Wendong and Wu, Yang and Yu, Qifan and Gao, Minghe and Miao, Bingchen and Zhang, Zhenkui and Pan, Kaihang and Li, Yunfei and Li, Mengze and Ji, Wei and others},
  booktitle={International Conference on Machine Learning},
  pages={5725--5748},
  year={2025},
  organization={PMLR}
}
}

\clearpage
\appendix

\clearpage
\setcounter{page}{1}
\maketitlesupplementary

\section*{Overview}

In this supplementary material, we present:

\begin{itemize}
    \item Formal definition of the \textbf{SCALE} framework as a POMDP, along with the self-check and iterative update algorithm and exploration-specific parameters, in Section~\ref{sec:scale-formulation}, Section~\ref{sec:scale-algorithm} and Section~\ref{sec:explore parameters}.
    \item Detailed prompt designs for the \textit{Selector}, \textit{Predictor}, and \textit{Judger} modules, in Section~\ref{sec:prompts}.
    \item Task construction prompts for the \textbf{SCALE-20k} dataset, as well as representative examples of single-step tasks, multi-step tasks, and page-level QA tasks from \textbf{SCALE-20k}, in Section~\ref{sec:scale-20k prompt} and Section~\ref{sec:scale-20k examples}.
    \item Comprehensive description of baselines, evaluation benchmarks, and standardized input protocols used in our experiments, in Section~\ref{sec:baselines}.
    \item Hardware configurations and training hyperparameters for fine-tuning \texttt{Qwen2.5-VL-7B-Instruct}, \texttt{InternVL2.5-8B} and \texttt{LLaVA‑NeXT‑8B}, in Section~\ref{sec:hardware}.
    \item Design of three dataset quality assessment metrics and evaluator prompts for reasoning quality and action optimality, in Section~\ref{sec:evaluation metrics} and Section~\ref{sec:evaluation prompts}.

\end{itemize}

\section{Details of SCALE Framework and SCALE-20k Dataset}

This section provides a comprehensive overview of the formal definition, algorithm, and implementation details of the \textbf{SCALE} framework, as well as the construction of the \textbf{SCALE-20k} dataset. It covers the POMDP formulation, the iterative self-check algorithm, prompt designs for each module, exploration-specific parameters, and representative dataset examples to facilitate reproducibility.

\subsection{Formulation of the SCALE}
\label{sec:scale-formulation}
We model the exploration process as a partially observable Markov decision process (POMDP), defined as a tuple $(S, A, O, T, R, \Omega)$, where $S$ denotes the set of environment states, $A$ is the action space, including actions such as \texttt{Click}, \texttt{Type}, \texttt{Hover} and \texttt{Scroll}, and $O$ represents the set of observations available to the agent. $T: S \times A \rightarrow S$ is the state transition function that models how the environment changes in response to an action, and $R$ is a binary function used to determine whether an exploration action is worth performing in a given state. $\Omega: S \rightarrow O$ is the observation function that maps the underlying state to the observable interface. This formulation captures the uncertainty and partial observability inherent in real-world web environments.

\subsection{Algorithm of the SCALE Framework}
\label{sec:scale-algorithm}
This is the specific algorithm of the SCALE method. We denote the environment at step $i$ as $S_i$, the observation as $O_i$, and the action as $a_i$, where $i \in [1, K]$, and $K$ is the number of interaction steps in a single self-update iteration. After $K$ steps, the agent undergoes one round of self-update. Let $j \in [1, N]$ denote the number of self-update rounds, with $N$ as the maximum number of updates allowed. The algorithm is shown in Algorithm~\ref{alg:self-check}.

\begin{algorithm}[t]
\caption{Self-Check and Iterative Update Process}
\label{alg:self-check}
\begin{algorithmic}[1]
\REQUIRE Initial policies $\pi^{sel}_1$, $\pi^{pre}_1$, fixed Judger, max update rounds $N$, steps per round $K$, initial state $S_1$
\FOR{$j = 1$ to $N$}
    \STATE Initialize buffer: $\mathcal{D}_j \gets \emptyset$
    \FOR{$i = 1$ to $K$}
        \STATE $O_i \gets \Omega(S_i)$
        \STATE $(a_i, r^{sel}_i) \gets \pi^{sel}_j(O_i)$
        \STATE \textit{repeat} $\gets$ \textbf{True}
        \WHILE{\textit{repeat}}
            \STATE $(p_i, r^{pre}_i) \gets \pi^{pre}_j(O_i, a_i, r^{sel}_i)$
            \STATE $S_{i+1} \gets T(S_i, a_i)$
            \STATE $O_{i+1} \gets \Omega(S_{i+1})$
            \STATE $j_i \gets \text{Judger}(O_i, O_{i+1}, a_i, p_i, r^{pre}_i)$
            \IF[Action not understood]{$j_i = 0$} 
                \STATE \textit{repeat} $\gets$ \textbf{False}
            \ELSE [Action understood, try again]
                \STATE $S_{i+1}$ is reverted to $S_i$
                \STATE $(a_i, r^{sel}_i) \gets \pi^{sel}_j(O_i, a_i, r^{sel}_i)$
            \ENDIF
        \ENDWHILE
        \STATE $(t_i, r^{t}_i) \gets \text{Judger}(O_i, O_{i+1}, a_i, r^{sel}_i)$
        \STATE $\mathcal{D}_j \gets \mathcal{D}_j \cup \{ (O_i, a_i, r^{sel}_i, t_i, r^t_i) \}$
    \ENDFOR
    \STATE $\pi^{sel}_{j+1} \gets \text{SFT}(\pi^{sel}_j, \mathcal{D}_j)$
    \STATE $\pi^{pre}_{j+1} \gets \text{SFT}(\pi^{pre}_j, \mathcal{D}_j)$
    \STATE $S_1 \gets S_{K}$
\ENDFOR
\end{algorithmic}
\end{algorithm}

\subsection{Exploration Specific Parameters}
\label{sec:explore parameters}
We define several key parameters in the \textbf{SCALE-Hop} module to support effective exploration and verification in web environments.

\noindent\textbf{Node Differentiation.}
To determine whether two nodes represent different web states, we jointly consider the \textbf{URL} and the visual similarity between observations. Specifically, we compute the \textbf{Structural Similarity Index Measure (SSIM)} between the two screenshots. A new node is created only if the URL differs or the SSIM value is below a strict threshold of 0.99, ensuring that visually similar pages are not redundantly revisited.

\noindent\textbf{Verification-Guided Backtracking.}
To avoid infinite loops in local exploration, we introduce a self-verification mechanism. If the agent remains within the same node for more than 10 consecutive steps without successful progression, it triggers Verification-Guided Backtracking. During this process, the system generates 5 candidate verification actions to determine whether the current state has been misunderstood or overlooked due to subtle changes in the environment. When there is link fluctuation or the exploration becomes stuck and difficult to continue, the self-verification mechanism is proactively activated to trigger verification-guided backtracking and switch to another node.

\subsection{Prompts Used in SCALE}
\label{sec:prompts}
In this section, we provide the prompts used in \textit{Selector}, \textit{Predictor}, and \textit{Judger} in \textbf{SCALE}.

\noindent\textbf{\textit{Selector} Prompt}
\newtcolorbox{selpromptbox}[1][]{
  colback=gray!10!white,
  colframe=black,
  width=1.0\columnwidth,
  arc=2mm,
  boxrule=0.5pt,
  title=\textbf{Prompt for \textit{Selector}},
  fonttitle=\normalsize\bfseries,
  enhanced,
  breakable,
  sharp corners,
  #1
}

\begin{center}
\begin{selpromptbox}
You are an exploration system designed to discover unknown interactions and the resulting changes in the environment. Your task is to generate a \textbf{single exploratory action} based on the provided image. 

The action of exploration needs to be \textbf{meaningful}! The action you generate should be one whose functionality and outcome are \textbf{uncertain and unpredictable}.

Based on the input image, generate an exploratory action. The generated action must satisfy the following requirements:

\begin{enumerate}[label=\textbf{\arabic*.}, leftmargin=1.5em]
    \item \textbf{The action command should start with action: followed by a concise command.} 
    (for example, \texttt{action: click [<insert item number in picture>]}, \texttt{type [<insert item number in picture>][<typing text>]}, \texttt{action: hover [<insert item number in picture>]}, \texttt{action: scroll [<down or up>]}).

    \item \textbf{The action command should be a simple command without any extra explanation.}

    \item \textbf{Immediately following the action command, provide a reason starting with \texttt{reason:}} 
    that explains why you chose this action and why its effect is unknown, requiring exploration.

    \item \textbf{The only possible actions you can generate are:} \texttt{scroll}, \texttt{click}, \texttt{hover}, \texttt{type}, or \texttt{fill}. 
    When you use \texttt{type} or \texttt{fill} action you must provide the specific element in the image and the fill content. 
    For example: \texttt{type [1][chips]}.

    \item \textbf{Your output must include both the action and reason parts}, separated by a newline, exactly in the following format:

{\ttfamily
action: <insert action>

reason: <insert why you choose this action, and why this action will lead to unknown interactions>
}

\end{enumerate}

In any given instance, you should generate only one action and its corresponding reason. 
If it hasn't been generated before, try \texttt{action: scroll [down]} to explore more content.

\end{selpromptbox}
\end{center}

\noindent\textbf{\textit{Predictor} Prompt}

\newtcolorbox{prepromptbox}[1][]{
  colback=gray!10!white,
  colframe=black,
  width=1.0\columnwidth,
  arc=2mm,
  boxrule=0.5pt,
  title=\textbf{Prompt for \textit{Predictor}},
  fonttitle=\normalsize\bfseries,
  enhanced,
  breakable,
  sharp corners,
  #1
}

\begin{center}
\begin{prepromptbox}

You are a prediction system designed to forecast the result of an action within a given environment. Your task is to \textbf{predict what will happen after the provided action is performed in the environment represented by the image.}\\

The result should be described in detail, focusing on the changes in the environment. 

Please predict the result after the action in the picture given. The action is as follows: \textbf{[ACTION]}

Describe the changes in the environment after the action is performed, focusing on the details of the transformation. If you think this action cannot change the environment, please answer accordingly that this action cannot make any changes.

The description should start with \texttt{describe:}, followed by a detailed account of the changes that might occur. After that, provide a reason for the prediction starting with \texttt{reason:}.\\

For example:

{\ttfamily
describe: <insert detailedly describe the changes in the environment after the action is performed, including specific details. The description must include changes in page content, interface elements, functionality, etc.>

reason: <insert reason why the described changes are expected, based on interface design rules, common user behavior patterns, or situational reasoning. You can mention page layout, interface interaction rules, or typical user actions and expectations.>
}

\end{prepromptbox}
\end{center}

\noindent\textbf{\textit{Judger} Prompt}

\newtcolorbox{judpromptbox}[1][]{
  colback=gray!10!white,
  colframe=black,
  width=1.0\columnwidth,
  arc=2mm,
  boxrule=0.5pt,
  title=\textbf{Prompt for \textit{Judger}},
  fonttitle=\normalsize\bfseries,
  enhanced,
  breakable,
  sharp corners,
  #1
}

\begin{center}
\begin{judpromptbox}

\textbf{\textit{System Prompt}}

You are a judgment system designed to verify whether the predicted result matches the actual outcome after the action is performed. Your task is to evaluate whether the prediction described matches the environment change shown in the picture. The first picture given is the original image before any action is performed, the second picture the image with labeled elements to assist in understanding the action, and the third picture shows the environment after the action has been performed. Your job is to determine if the prediction matches the changes between the first picture and the third picture after the action was performed. \\

If the prediction matches the outcome, return \texttt{judge: TRUE}. If the prediction does not match, or uses vague descriptions like a new page appears or the layout changes, return \texttt{judge: FALSE}, and provide a correct description of the action's result starting with \texttt{true describe:}.

In \texttt{true describe}, you should list the most significant modifications and then describe every small detail that has changed (e.g., text content, button positions, colors, UI layout, icons, images, main function changes, etc.).

\textbf{!!! Avoid vague descriptions like a `new page appears' or `the layout changes'. Instead, describe exactly what is different.}\\

Please output your response in the following format:
\begin{itemize}
    \item \texttt{judge: TRUE} if the prediction matches the outcome.
    \item If it does not match, \texttt{judge: FALSE} and provide the correct description starting with \texttt{true describe:}, then give the true judgment reason starting with \texttt{true reason:}
\end{itemize}

\textbf{Example 1:}
\begin{flushleft}
\ttfamily
judge: TRUE
\end{flushleft}

\textbf{Example 2:}
\begin{flushleft}
\ttfamily
judge: FALSE
true describe: <insert describe detailedly the change between the first picture and the third picture, which is the change after the action.>
true reason: <insert analyze the real reason why the first picture will produce the third picture result after this action.(the action label is shown in the second picture)> 
\end{flushleft}

\textbf{\\User Prompt}

The action performed is described as: \textbf{[ACTION]}

The predicted result is described as: \textbf{[PREDICTION]}

Determine whether the predicted result matches the actual outcome in the third image. If it matches and describes detailedly, respond with \texttt{judge: TRUE}.

If it does \textbf{NOT} match or does \textbf{NOT} describe detailedly, respond with \texttt{judge: FALSE}, and provide a correct and detailed description of the action's result starting with \texttt{true describe:}.

\textbf{Important:} Your description must be detailed and specific. If the predicted result is inaccurate or the description is too vague, it should be considered incorrect.

When answering, do \textbf{not} mention `first image' or `third image', but rather the original environment and the environment after the action.

Provide enough detail to explain what exactly changed in the environment after the action was performed, and make sure to describe any visual or functional differences in the image.\\

\textbf{Note:} The third image shows no environmental changes, meaning no alterations occurred after performing the action. \textit{(Optional hint when \texttt{same==1})}

\end{judpromptbox}
\end{center}

\subsection{SCALE-20k Prompt}
\label{sec:scale-20k prompt}
In this section, we introduce the construction process of the \textbf{SCALE-20k} dataset. Specifically, we detail the prompt formulations used to generate three types of supervision signals—single-step tasks, multi-step tasks, and page-level question answering (QA) pairs. The \textbf{SCALE-20k} dataset includes data collected from a total of 19 websites, four from VisualWebArena and fifteen from WebVoyager. 

\noindent\textbf{Action description generation Prompt}
\newtcolorbox{despromptbox}[1][]{
  colback=gray!10!white,
  colframe=black,
  width=1.0\columnwidth,
  arc=2mm,
  boxrule=0.5pt,
  title=\textbf{Prompt for Action Description},
  fonttitle=\normalsize\bfseries,
  enhanced,
  breakable,
  sharp corners,
  #1
}

\begin{center}
\begin{despromptbox}
The image shows the current state of the webpage interface.
The corresponding action is: \textbf{[ACTION]}

Your task is to generate a simple, accurate, and direct one-sentence description of the operation implied by this action.

\begin{itemize}
    \item Do not include the action syntax like \texttt{click [id]} or \texttt{type [id] [content]} in your response.
    \item Do not use words like \textit{label}, \textit{id}, or technical terms — just describe what the user would do.
    \item \textbf{IMPORTANT:} The description must align with the action type; for example, do not describe a \texttt{click} action using the word \textit{hover} or other unrelated verbs.
    \item Only output the natural language description. Do not include \texttt{Action:} or any other prefixes in your answer.
\end{itemize}

\textbf{\\Examples (input → output):}
\begin{itemize}
    \item Action: \texttt{click [12]} → Click the `Event' button.
    \item Action: \texttt{type [6] [xxx]} → Type xxx in the name field.
    \item Action: \texttt{scroll [down]} → Scroll the page down to view more listings.
\end{itemize}
\end{despromptbox}
\end{center}

\noindent\textbf{Single-Step Task Construction Prompt}
\newtcolorbox{singpromptbox}[1][]{
  colback=gray!10!white,
  colframe=black,
  width=1.0\columnwidth,
  arc=2mm,
  boxrule=0.5pt,
  title=\textbf{Prompt for Single-Step Task Construction},
  fonttitle=\normalsize\bfseries,
  enhanced,
  breakable,
  sharp corners,
  #1
}

\begin{center}
\begin{singpromptbox}

The provided image shows the current status of the website environment.
Please generate higher-level tasks based on this action \textbf{[ACTION]} and the object \textbf{[ACTION DESCRIBE]} it operates on.

Do not refer to its border — it is marked only for identifying the location. Focus on the content of the item itself.\\

\textbf{Examples:}
\begin{itemize}
    \item Task: Buy me the T-shirt with darkest color.
    \item Task: Add the cheapest item to my wish list.
    \item Task: Can you search watch for me.
    \item Task: Scroll down to load more products in the current category.
\end{itemize}

The task should be a simple instruction that can be executed in the current environment.

\begin{itemize}
    \item \textbf{DO NOT} generate literal tasks like “Click [object]” or “Do xxx”.
    \item Combine the object's function and the action's result to form a meaningful user goal.
\end{itemize}

You should only return the task and nothing else, following the format below:

\texttt{Task: <insert the generated task here>}

\end{singpromptbox}
\end{center}

\noindent\textbf{Multi-Step Task Construction Prompt}

\newtcolorbox{multipromptbox}[1][]{
  colback=gray!10!white,
  colframe=black,
  width=1.0\columnwidth,
  arc=2mm,
  boxrule=0.5pt,
  title=\textbf{Prompt for Multi-Step Task Construction (Three Steps)},
  fonttitle=\normalsize\bfseries,
  enhanced,
  breakable,
  sharp corners,
  #1
}

\begin{center}
\begin{multipromptbox}

You are given four interface screenshots representing a user workflow:

\begin{enumerate}
    \item \textbf{Initial State (First Image)}: The environment before any action.
    \item \textbf{After Action 1 (Second Image)}: The environment after executing \textbf{Action 1}.
    \item \textbf{After Action 2 (Third Image)}: The environment after executing \textbf{Action 2}.
    \item \textbf{After Action 3 (Fourth Image)}: The environment after executing \textbf{Action 3}, which should represent the final task completion state.
\end{enumerate}

The three actions performed are:

\begin{itemize}
    \item Action 1: \textbf{[ACTION 1]}
    \item Action 2: \textbf{[ACTION 2]}
    \item Action 3: \textbf{[ACTION 3]}
\end{itemize}

You are also provided with abstract descriptions of the visual and contextual features of the targets for each action:

\begin{itemize}
    \item For Action 1: \textbf{[ACTION DESCRIBE 1]}
    \item For Action 2: \textbf{[ACTION DESCRIBE 2]}
    \item For Action 3: \textbf{[ACTION DESCRIBE 3]}
\end{itemize}

These describe what the target looks like or its purpose (e.g., position, layout, category), but not its exact label or wording.

\textbf{Your task} is to generate a realistic and specific user goal that is fully completed by executing the above three actions \textbf{in order}.

\begin{itemize}
    \item The task should reflect a natural user intent — not a technical UI action description.
    \item Avoid phrasing the task as a sequence of steps (e.g., `first do A, then do B, then do C').
    \item Keep the task concise, meaningful, and achievable with the provided actions.
    \item The generated task must have been fully completed in the fourth image.
\end{itemize}

\textbf{\\Examples of valid tasks:}

\begin{itemize}
    \item Add a discounted smartphone to the cart by entering the smartphone category and selecting an item under \$300.
    \item Submit a review for a recently ordered vacuum cleaner by opening its product page and filling out the review form.
    \item Find a blue sofa listed in the top-rated furniture section and initiate a purchase.
    \item View the seller's contact options for a kitchen blender by navigating to the product details page.
    \item Open the support chat window to get help with a delayed shoe delivery.
\end{itemize}

Now, only return the final task using the following format:

\texttt{Task: <insert your generated task here>}

\end{multipromptbox}
\end{center}

\noindent\textbf{Page QA Construction Prompt}

\newtcolorbox{QApromptbox}[1][]{
  colback=gray!10!white,
  colframe=black,
  width=1.0\columnwidth,
  arc=2mm,
  boxrule=0.5pt,
  title=\textbf{Prompt for Page QA Construction},
  fonttitle=\normalsize\bfseries,
  enhanced,
  breakable,
  sharp corners,
  #1
}

\begin{center}
\begin{QApromptbox}

You are a useful task generation assistant. I will provide you with a screenshot of a web page and you will generate a question and answer pair based on the content of this screenshot.\\

\textbf{Questions} must strictly align with the content of the image and relate directly to it. Do not generate questions about elements outside the image.

\textbf{Answers} to the questions must be strictly derived from the visual information provided in the image. You can't fabricate the answer yourself.

The generated answer should be concise and direct. It should not be a long text answer. It is best to provide a brief answer with numbers or a price range.\\

\textbf{Example 1:}
\begin{itemize}
    \item Question: What is the price range of the products on the first line?
    \item Answer: \$1.00 -- \$600.00
\end{itemize}

\textbf{Example 2:}
\begin{itemize}
    \item Question: Tell me the price of the yellow item in the first row.
    \item Answer: \$10.00
\end{itemize}

\textbf{Example 3:}
\begin{itemize}
    \item Question: What is the length of the product in inches?
    \item Answer: 40.00
\end{itemize}

\textbf{Example 4:}
\begin{itemize}
    \item Question: What is the price of the darkest product in this page?
    \item Answer: \$12.00
\end{itemize}

\textbf{Example 5:}
\begin{itemize}
    \item Question: Tell me the price of the product with owl in the page.
    \item Answer: \$45.00
\end{itemize}

\textbf{IMPORTANT:} Only one question–answer pair can be generated at a time. The generated pair must be relevant to the given image, and the answer should be short and meaningful.\\

You should follow the output format exactly, with nothing else:

\texttt{Question: <Your generated task>}

\texttt{Answer: <Your generated answer of the question>}

\end{QApromptbox}
\end{center}

\noindent\textbf{Think Process Prompt}
In task-oriented web environments, reasoning about the next action based on the current observation is essential for agent decision-making. We design think process prompts to help the model infer the immediate intention behind the first action in a given task context.

While different types of tasks may involve different reasoning strategies depending on task structure and action dependencies, we unify the prompt format across examples with slight variations depending on action ordering. Here, we present the think process prompt tailored for a two-step task, where the reasoning focuses on initiating the \textbf{first action}.

\newtcolorbox{thinkpromptbox}[1][]{
  colback=gray!10!white,
  colframe=black,
  width=1.0\columnwidth,
  arc=2mm,
  boxrule=0.5pt,
  title=\textbf{Prompt for Think Process Construction},
  fonttitle=\normalsize\bfseries,
  enhanced,
  breakable,
  sharp corners,
  #1
}

\begin{center}
\begin{thinkpromptbox}

\textbf{System Prompt}

You are an assistant that generates simple and direct reasoning based on a task and current web environment.
Only focus on the present situation and what needs to be done first, without predicting future outcomes.

Use `I' as the subject, and keep your answer to one short paragraph (1–2 sentences max).
Write in a natural, direct tone. Avoid any lists, explanations, or formatting.\\
Avoid using phrases like ``step-by-step'', ``first'', ``second'', etc.

\textbf{Examples:}
\begin{itemize}
    \item Let's think step-by-step. I need to check a comment, and the post is right here, so I should go into it.
    \item Let's think step-by-step. To add a product to the shopping cart, I need to navigate to the catalog or product section. The ``CATALOG" link is available with ID [1234].\\
\end{itemize}

\textbf{User Prompt}

The first image shows the current website environment.

The second image shows the result after executing \textbf{[ACTION 1]}.\\
The task is: \textbf{[TASK]}.\\
The related actions are \textbf{[ACTION 1]} and \textbf{[ACTION 2]}.\\
In this initial situation, you need to think about executing the first action \textbf{[ACTION 1]} first.

Generate a short reasoning based only on the environment and the task.

\textbf{Important:}
\begin{itemize}
    \item Keep it very short and natural.
    \item Do not describe outcomes.
    \item Do not use formatting or list-style language.
    \item Output just one short paragraph.
\end{itemize}

\end{thinkpromptbox}
\end{center}

\subsection{SCALE-20k Examples.} 
\label{sec:scale-20k examples}
In this section, we provide specific SCALE-20k examples, including examples of single-step tasks, multi-step tasks, and page QA tasks. 

\newtcolorbox{singexamplebox}[1][]{
  colback=gray!10!white,
  colframe=black,
  width=1.0\columnwidth,
  arc=2mm,
  boxrule=0.5pt,
  title=\textbf{Example of Single-Step Tasks},
  fonttitle=\normalsize\bfseries,
  enhanced,
  breakable,
  sharp corners,
  #1
}

\begin{singexamplebox}

\textbf{User Input:}

\textbf{Image Observation:}

\begin{center}
\includegraphics[width=0.95\linewidth]{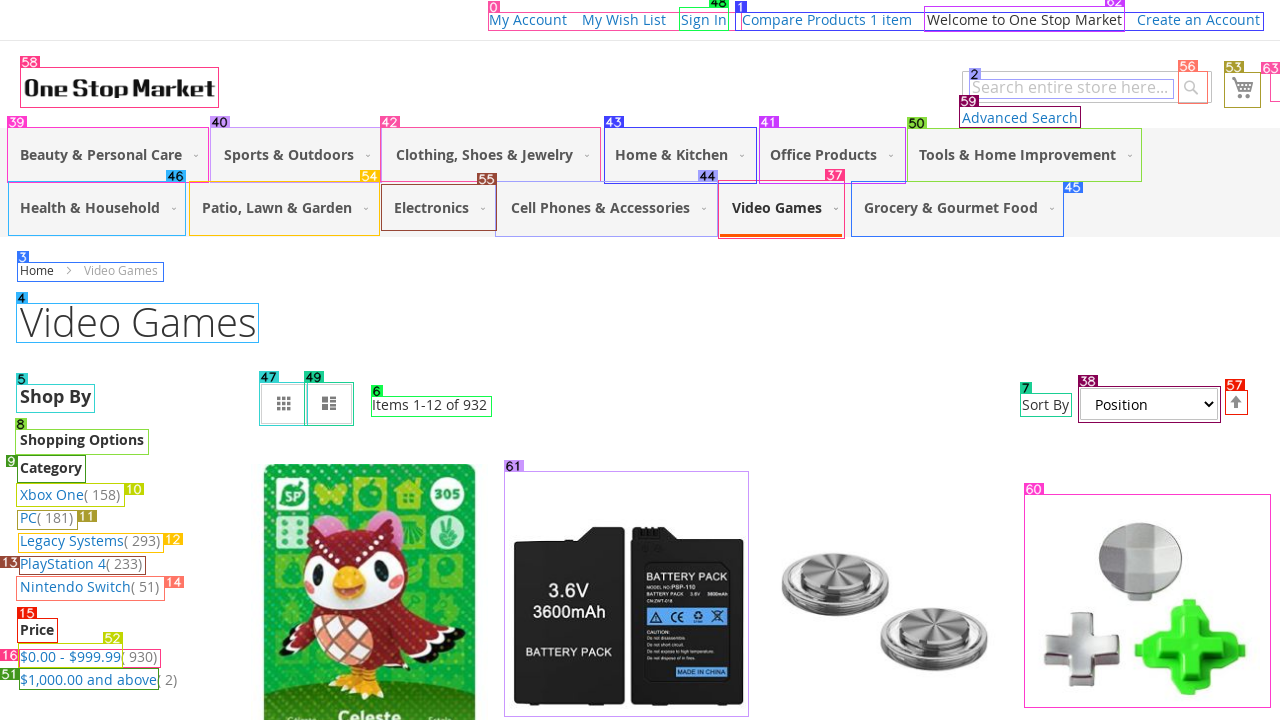} 
\end{center}

\textbf{Task Description:} 

You are an intelligent agent completing web-based tasks. Based on the user's objective (i.e. instruction), current interface information (i.e. screenshot and its corresponding accessibility tree), and action history, determine the next action.

\textbf{Available Actions:}
\begin{itemize}
  \item \texttt{click [id]}: This action clicks on an element with a specific id on the webpage.
  \item \texttt{type [id] [content] [press\_enter\_after=0|1]}: Use this to type the content into the field with id. By default, the ``Enter" key is pressed after typing unless press\_enter\_after is set to 0.
  \item \texttt{hover [id]}: Hover over an element with id.
  \item \texttt{press [key\_comb]}: Simulates the pressing of a key combination on the keyboard (e.g., Ctrl+v).
  \item \texttt{scroll [direction=down|up]}: Scroll the page up or down.
  \item \texttt{new\_tab}: Open a new, empty browser tab.
  \item \texttt{tab\_focus [tab\_index]}: Switch the browser's focus to a specific tab using its index.
  \item \texttt{close\_tab}: Close the currently active tab.
  \item \texttt{goto [url]}: Navigate to a specific URL.
  \item \texttt{go\_back}: Navigate to the previously viewed page.
  \item \texttt{go\_forward}: Navigate to the next page (if a previous `go\_back' action was performed).
  \item \texttt{stop [answer]}: Issue this action when you believe the task is complete. If the objective is to find a text-based answer, provide the answer in the bracket. If you believe the task is impossible to complete, provide the answer as ``N/A" in the bracket.
\end{itemize}

\textbf{Output Format:} 

First, generate the reasoning process for the action. Then, generate the action in the correct format. Start with a "In summary, the next action I will perform is" phrase, followed by action inside \texttt{\textasciigrave\textasciigrave\textasciigrave\textasciigrave\textasciigrave\textasciigrave}. For example, Let's think step-by-step. To add a product to the shopping cart, I need to navigate to the catalog or product section. The ``CATALOG" link is available with ID [1234]. In summary, the next action I will perform is \textasciigrave\textasciigrave\textasciigrave click [1234]\textasciigrave\textasciigrave\textasciigrave\\

\textbf{OBJECTIVE:} 

Navigate back to the homepage to explore more categories for shopping.

\textbf{PREVIOUS ACTION:} 

\textbf{IMAGES: (1)}\\

\textbf{Assistant Output:}

To get back to the homepage and explore more categories, I need to find a way to return to the main page. I can see the One Stop Market logo, which is a common website feature for returning to the homepage, so I should click on it to navigate back. In summary, the next action I will perform is \texttt{\textasciigrave\textasciigrave\textasciigrave click [58]\textasciigrave\textasciigrave\textasciigrave}

\end{singexamplebox}

\newtcolorbox{multiexamplebox}[1][]{
  colback=gray!10!white,
  colframe=black,
  width=1.0\columnwidth,
  arc=2mm,
  boxrule=0.5pt,
  title=\textbf{Example of Multi-Step Tasks},
  fonttitle=\normalsize\bfseries,
  enhanced,
  breakable,
  sharp corners,
  #1
}

\begin{multiexamplebox}

\textbf{STEP 1:}

\textbf{User Input:}

\textbf{Image Observation:}

\begin{center}
\includegraphics[width=0.95\linewidth]{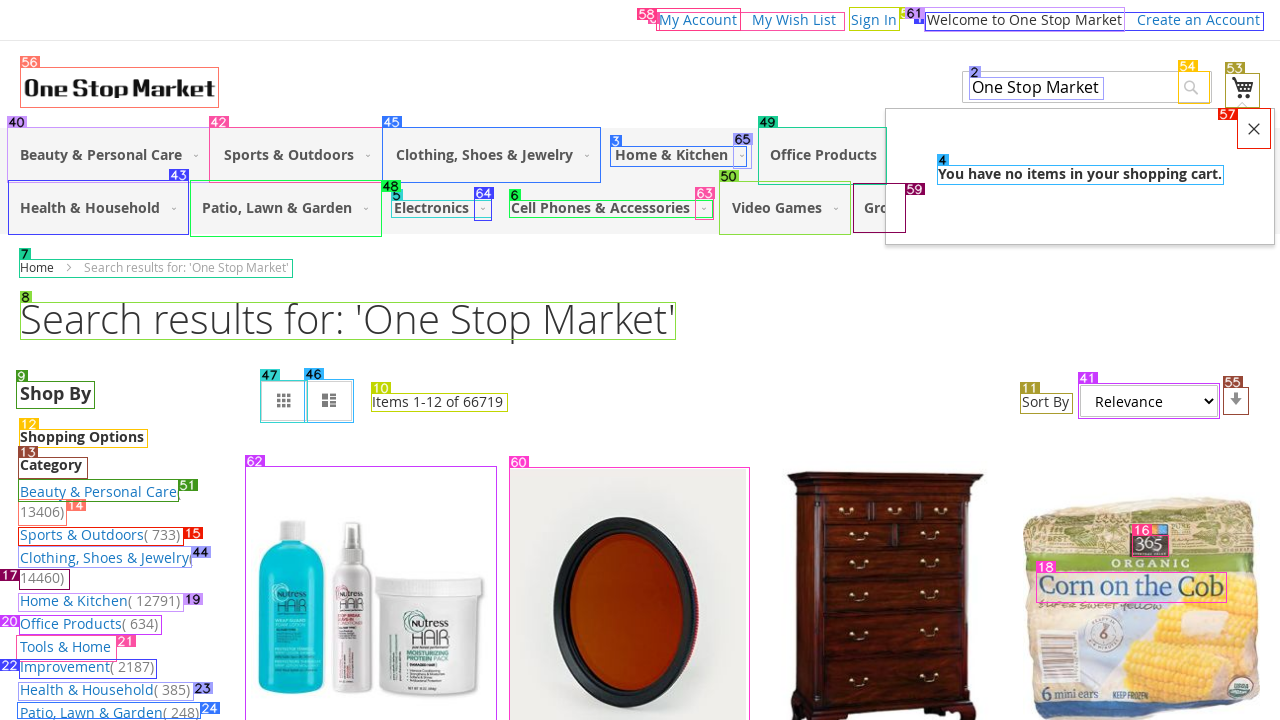} 
\end{center}

\textbf{Task Description:} 

You are an intelligent agent completing web-based tasks. Based on the user's objective (i.e. instruction), current interface information (i.e. screenshot and its corresponding accessibility tree), and action history, determine the next action.

\textbf{Available Actions:}
\begin{itemize}
  \item \texttt{click [id]}: This action clicks on an element with a specific id on the webpage.
  \item \texttt{type [id] [content] [press\_enter\_after=0|1]}: Use this to type the content into the field with id. By default, the ``Enter" key is pressed after typing unless press\_enter\_after is set to 0.
  \item \texttt{hover [id]}: Hover over an element with id.
  \item \texttt{press [key\_comb]}: Simulates the pressing of a key combination on the keyboard (e.g., Ctrl+v).
  \item \texttt{scroll [direction=down|up]}: Scroll the page up or down.
  \item \texttt{new\_tab}: Open a new, empty browser tab.
  \item \texttt{tab\_focus [tab\_index]}: Switch the browser's focus to a specific tab using its index.
  \item \texttt{close\_tab}: Close the currently active tab.
  \item \texttt{goto [url]}: Navigate to a specific URL.
  \item \texttt{go\_back}: Navigate to the previously viewed page.
  \item \texttt{go\_forward}: Navigate to the next page (if a previous `go\_back' action was performed).
  \item \texttt{stop [answer]}: Issue this action when you believe the task is complete. If the objective is to find a text-based answer, provide the answer in the bracket. If you believe the task is impossible to complete, provide the answer as ``N/A" in the bracket.
\end{itemize}

\textbf{Output Format:} 

First, generate the reasoning process for the action. Then, generate the action in the correct format. Start with a "In summary, the next action I will perform is" phrase, followed by action inside \texttt{\textasciigrave\textasciigrave\textasciigrave\textasciigrave\textasciigrave\textasciigrave}. For example, Let's think step-by-step. To add a product to the shopping cart, I need to navigate to the catalog or product section. The ``CATALOG" link is available with ID [1234]. In summary, the next action I will perform is \textasciigrave\textasciigrave\textasciigrave click [1234]\textasciigrave\textasciigrave\textasciigrave\\

\textbf{OBJECTIVE:} 

Browse the available video game accessories under \$999.99.

\textbf{PREVIOUS ACTION:} 

\textbf{IMAGES: (1)}\\

\textbf{Assistant Output:}

I need to click on the One Stop Market logo marked with [56] to navigate to the homepage. In summary, the next action I will perform is \texttt{\textasciigrave\textasciigrave\textasciigrave click [56]\textasciigrave\textasciigrave\textasciigrave}\\

\textbf{STEP 2:}

\textbf{User Input:}

\textbf{Image Observation:}

\begin{center}
\includegraphics[width=0.95\linewidth]{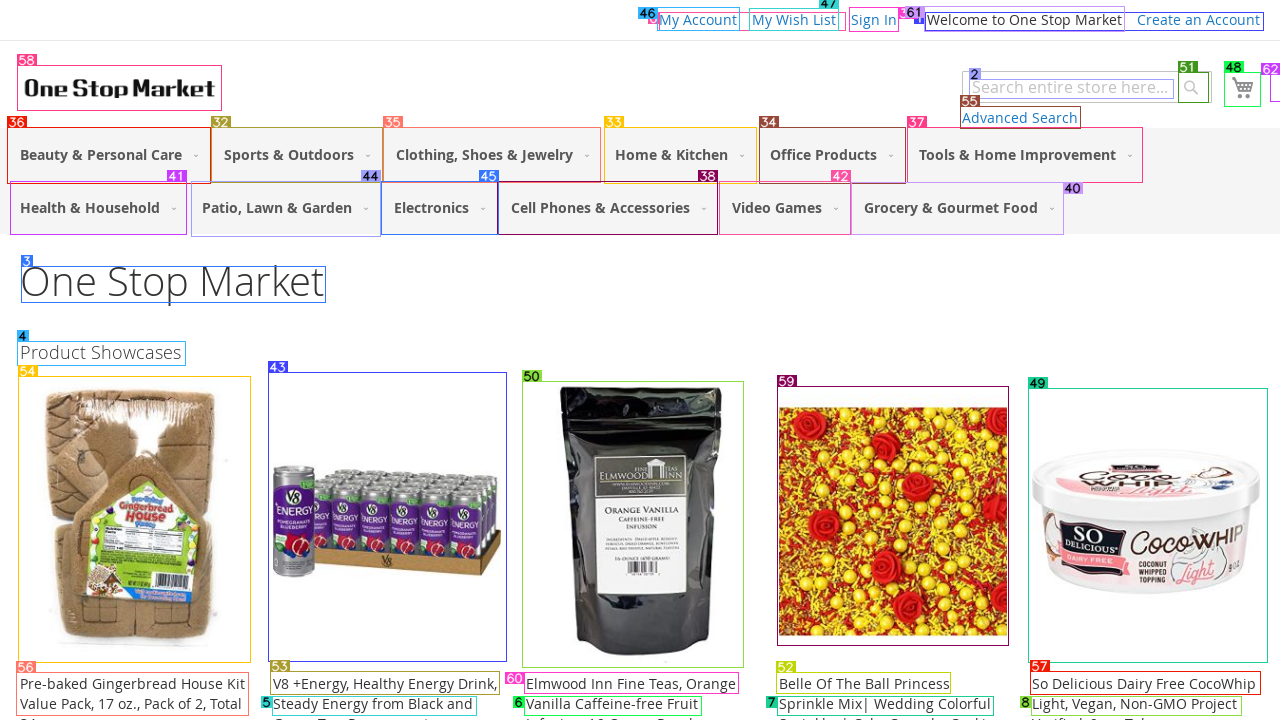} 
\end{center}

\textbf{Task Description:} 

You are an intelligent agent completing web-based tasks. Based on the user's objective (i.e. instruction), current interface information (i.e. screenshot and its corresponding accessibility tree), and action history, determine the next action.

\textbf{Available Actions:}
\begin{itemize}
  \item \texttt{click [id]}: This action clicks on an element with a specific id on the webpage.
  \item \texttt{type [id] [content] [press\_enter\_after=0|1]}: Use this to type the content into the field with id. By default, the ``Enter" key is pressed after typing unless press\_enter\_after is set to 0.
  \item \texttt{hover [id]}: Hover over an element with id.
  \item \texttt{press [key\_comb]}: Simulates the pressing of a key combination on the keyboard (e.g., Ctrl+v).
  \item \texttt{scroll [direction=down|up]}: Scroll the page up or down.
  \item \texttt{new\_tab}: Open a new, empty browser tab.
  \item \texttt{tab\_focus [tab\_index]}: Switch the browser's focus to a specific tab using its index.
  \item \texttt{close\_tab}: Close the currently active tab.
  \item \texttt{goto [url]}: Navigate to a specific URL.
  \item \texttt{go\_back}: Navigate to the previously viewed page.
  \item \texttt{go\_forward}: Navigate to the next page (if a previous `go\_back' action was performed).
  \item \texttt{stop [answer]}: Issue this action when you believe the task is complete. If the objective is to find a text-based answer, provide the answer in the bracket. If you believe the task is impossible to complete, provide the answer as ``N/A" in the bracket.
\end{itemize}

\textbf{Output Format:} 

First, generate the reasoning process for the action. Then, generate the action in the correct format. Start with a "In summary, the next action I will perform is" phrase, followed by action inside \texttt{\textasciigrave\textasciigrave\textasciigrave\textasciigrave\textasciigrave\textasciigrave}. For example, Let's think step-by-step. To add a product to the shopping cart, I need to navigate to the catalog or product section. The ``CATALOG" link is available with ID [1234]. In summary, the next action I will perform is \textasciigrave\textasciigrave\textasciigrave click [1234]\textasciigrave\textasciigrave\textasciigrave\\

\textbf{OBJECTIVE:} 

Browse the available video game accessories under \$999.99.

\textbf{PREVIOUS ACTION:} 

click [56]

\textbf{IMAGES: (1)}\\

\textbf{Assistant Output:}

Let's think step-by-step. To browse video game accessories, I need to navigate to the Video Games section, which is accessible through the Video Games category link with ID [42]. In summary, the next action I will perform is \texttt{\textasciigrave\textasciigrave\textasciigrave click [42]\textasciigrave\textasciigrave\textasciigrave}\\

\textbf{STEP 3:}

\textbf{User Input:}

\textbf{Image Observation:}

\begin{center}
\includegraphics[width=0.95\linewidth]{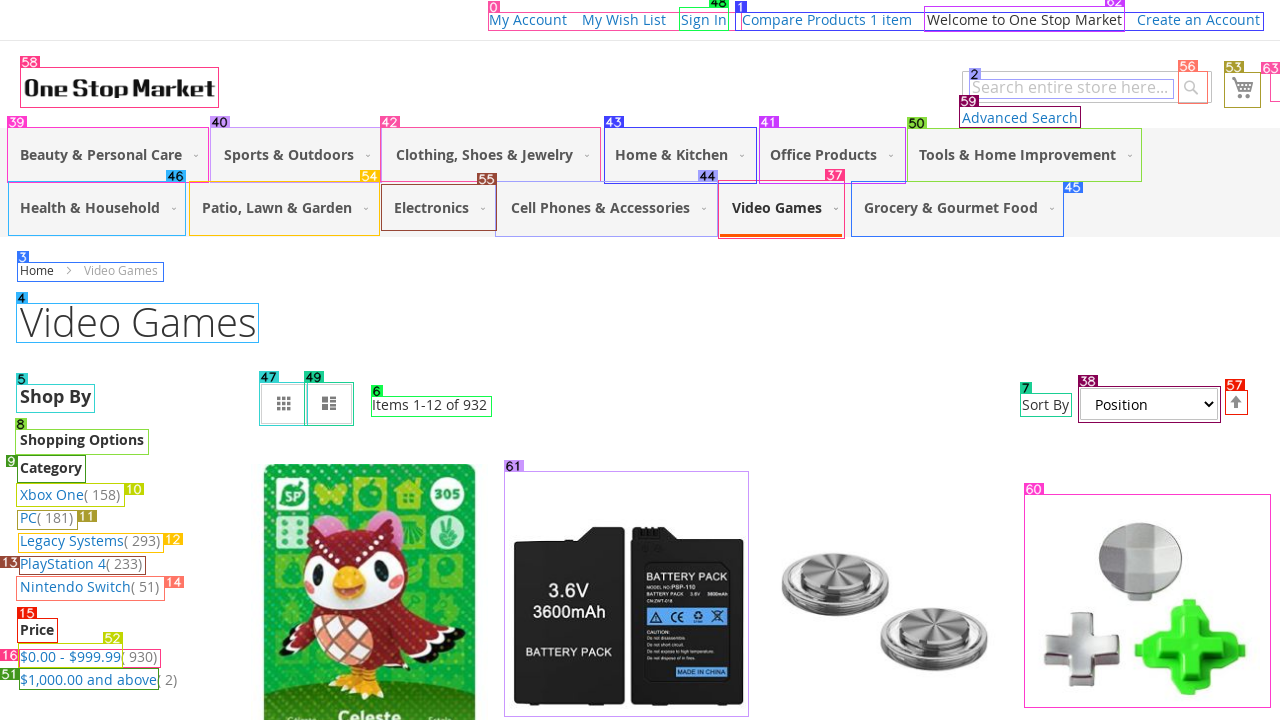} 
\end{center}

\textbf{Task Description:} 

You are an intelligent agent completing web-based tasks. Based on the user's objective (i.e. instruction), current interface information (i.e. screenshot and its corresponding accessibility tree), and action history, determine the next action.

\textbf{Available Actions:}
\begin{itemize}
  \item \texttt{click [id]}: This action clicks on an element with a specific id on the webpage.
  \item \texttt{type [id] [content] [press\_enter\_after=0|1]}: Use this to type the content into the field with id. By default, the ``Enter" key is pressed after typing unless press\_enter\_after is set to 0.
  \item \texttt{hover [id]}: Hover over an element with id.
  \item \texttt{press [key\_comb]}: Simulates the pressing of a key combination on the keyboard (e.g., Ctrl+v).
  \item \texttt{scroll [direction=down|up]}: Scroll the page up or down.
  \item \texttt{new\_tab}: Open a new, empty browser tab.
  \item \texttt{tab\_focus [tab\_index]}: Switch the browser's focus to a specific tab using its index.
  \item \texttt{close\_tab}: Close the currently active tab.
  \item \texttt{goto [url]}: Navigate to a specific URL.
  \item \texttt{go\_back}: Navigate to the previously viewed page.
  \item \texttt{go\_forward}: Navigate to the next page (if a previous `go\_back' action was performed).
  \item \texttt{stop [answer]}: Issue this action when you believe the task is complete. If the objective is to find a text-based answer, provide the answer in the bracket. If you believe the task is impossible to complete, provide the answer as ``N/A" in the bracket.
\end{itemize}

\textbf{Output Format:} 

First, generate the reasoning process for the action. Then, generate the action in the correct format. Start with a "In summary, the next action I will perform is" phrase, followed by action inside \texttt{\textasciigrave\textasciigrave\textasciigrave\textasciigrave\textasciigrave\textasciigrave}. For example, Let's think step-by-step. To add a product to the shopping cart, I need to navigate to the catalog or product section. The ``CATALOG" link is available with ID [1234]. In summary, the next action I will perform is \textasciigrave\textasciigrave\textasciigrave click [1234]\textasciigrave\textasciigrave\textasciigrave\\

\textbf{OBJECTIVE:} 

Browse the available video game accessories under \$999.99.

\textbf{PREVIOUS ACTION:} 

click [56] click [42]

\textbf{IMAGES: (1)}\\

\textbf{Assistant Output:}

Let's think step-by-step. To browse video game accessories under \$999.99, I click on the price filter [16] that matches the set range. In summary, the next action I will perform is \texttt{\textasciigrave\textasciigrave\textasciigrave click [16]\textasciigrave\textasciigrave\textasciigrave}\\

\textbf{STEP 4:}

\textbf{User Input:}

\textbf{Image Observation:}

\begin{center}
\includegraphics[width=0.95\linewidth]{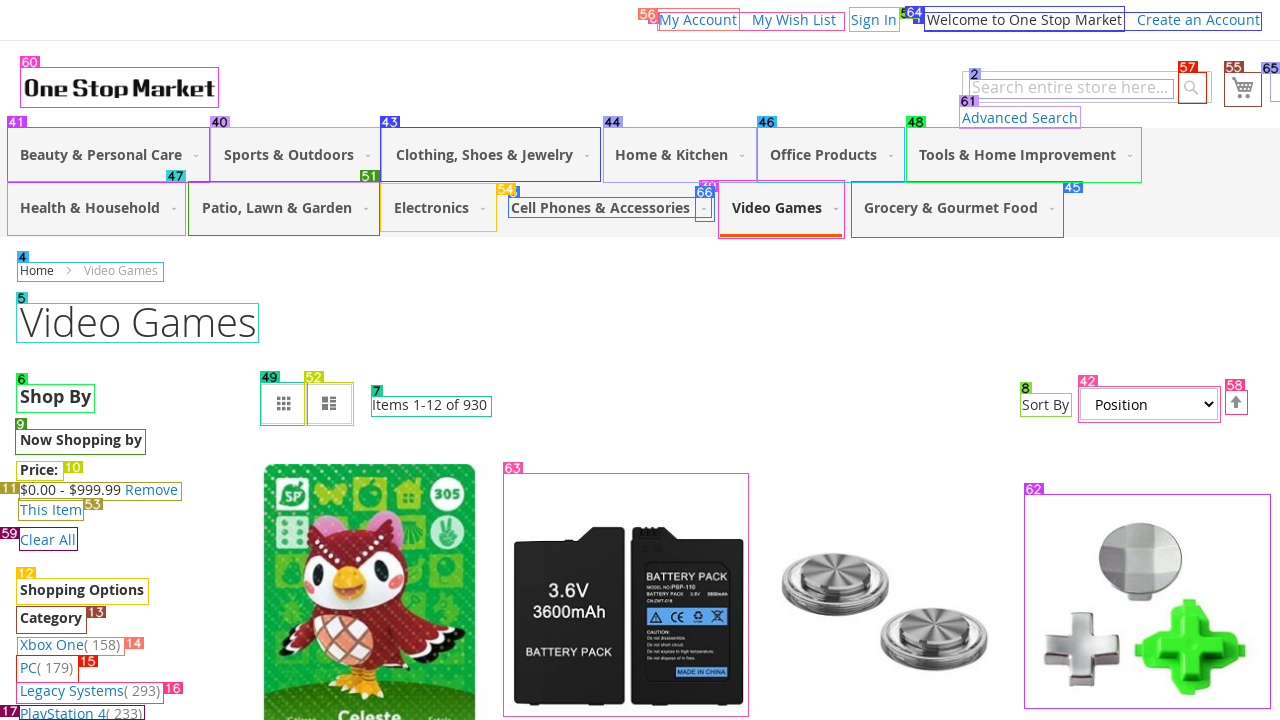} 
\end{center}

\textbf{Task Description:} 

You are an intelligent agent completing web-based tasks. Based on the user's objective (i.e. instruction), current interface information (i.e. screenshot and its corresponding accessibility tree), and action history, determine the next action.

\textbf{Available Actions:}
\begin{itemize}
  \item \texttt{click [id]}: This action clicks on an element with a specific id on the webpage.
  \item \texttt{type [id] [content] [press\_enter\_after=0|1]}: Use this to type the content into the field with id. By default, the ``Enter" key is pressed after typing unless press\_enter\_after is set to 0.
  \item \texttt{hover [id]}: Hover over an element with id.
  \item \texttt{press [key\_comb]}: Simulates the pressing of a key combination on the keyboard (e.g., Ctrl+v).
  \item \texttt{scroll [direction=down|up]}: Scroll the page up or down.
  \item \texttt{new\_tab}: Open a new, empty browser tab.
  \item \texttt{tab\_focus [tab\_index]}: Switch the browser's focus to a specific tab using its index.
  \item \texttt{close\_tab}: Close the currently active tab.
  \item \texttt{goto [url]}: Navigate to a specific URL.
  \item \texttt{go\_back}: Navigate to the previously viewed page.
  \item \texttt{go\_forward}: Navigate to the next page (if a previous `go\_back' action was performed).
  \item \texttt{stop [answer]}: Issue this action when you believe the task is complete. If the objective is to find a text-based answer, provide the answer in the bracket. If you believe the task is impossible to complete, provide the answer as ``N/A" in the bracket.
\end{itemize}

\textbf{Output Format:} 

First, generate the reasoning process for the action. Then, generate the action in the correct format. Start with a "In summary, the next action I will perform is" phrase, followed by action inside \texttt{\textasciigrave\textasciigrave\textasciigrave\textasciigrave\textasciigrave\textasciigrave}. For example, Let's think step-by-step. To add a product to the shopping cart, I need to navigate to the catalog or product section. The ``CATALOG" link is available with ID [1234]. In summary, the next action I will perform is \textasciigrave\textasciigrave\textasciigrave click [1234]\textasciigrave\textasciigrave\textasciigrave\\

\textbf{OBJECTIVE:} 

Browse the available video game accessories under \$999.99.

\textbf{PREVIOUS ACTION:} 

click [56] click [42] click [16]

\textbf{IMAGES: (1)}\\

\textbf{Assistant Output:}

I have displayed video game accessories filtered by the price range of \$0.00 - \$999.99. In summary, the next action I will perform is \texttt{\textasciigrave\textasciigrave\textasciigrave stop[I have browsed the available video game accessories under \$999.99]\textasciigrave\textasciigrave\textasciigrave}.
\end{multiexamplebox}

\newtcolorbox{QAexamplebox}[1][]{
  colback=gray!10!white,
  colframe=black,
  width=1.0\columnwidth,
  arc=2mm,
  boxrule=0.5pt,
  title=\textbf{Example of Page QA Tasks},
  fonttitle=\normalsize\bfseries,
  enhanced,
  breakable,
  sharp corners,
  #1
}

\begin{center}
\begin{QAexamplebox}

\textbf{User Input:}

\textbf{Image Observation:}

\begin{center}
\includegraphics[width=0.95\linewidth]{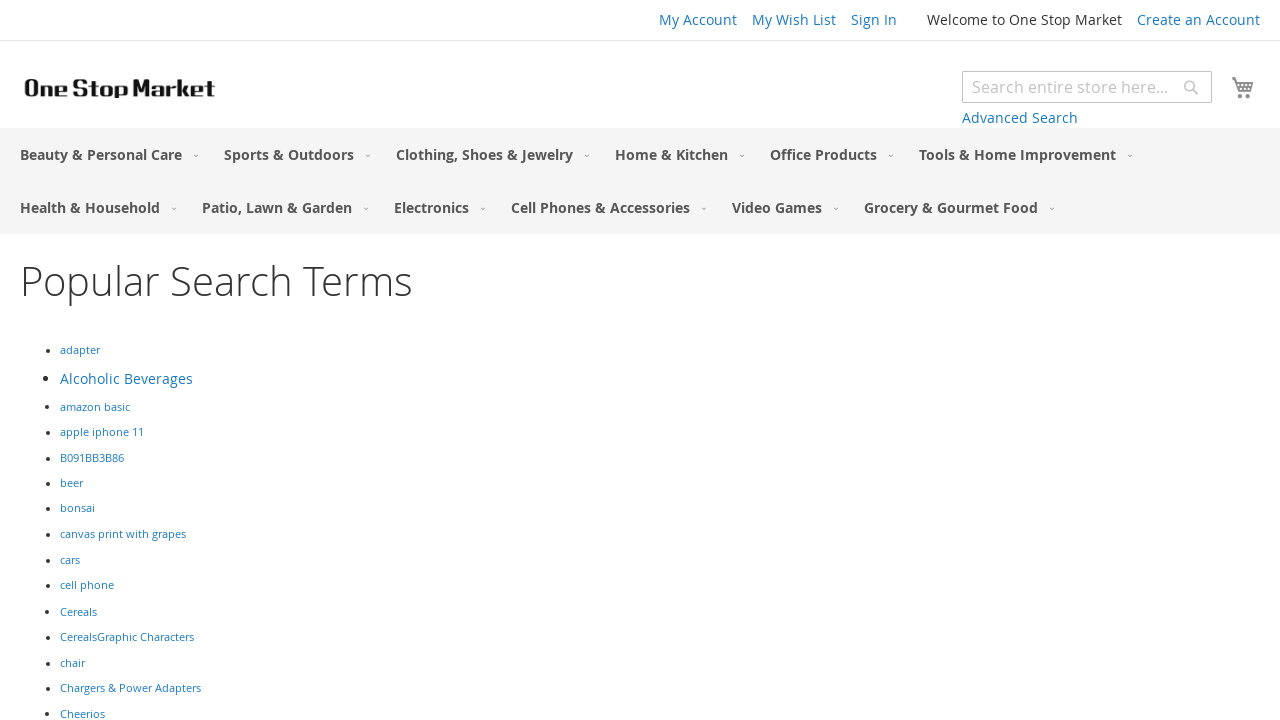} 
\end{center}

\textbf{Task Description:} 

You are an intelligent agent completing web-based tasks. Based on the user's objective (i.e. instruction), current interface information (i.e. screenshot and its corresponding accessibility tree), and action history, determine the next action.

\textbf{Available Actions:}
\begin{itemize}
  \item \texttt{click [id]}: This action clicks on an element with a specific id on the webpage.
  \item \texttt{type [id] [content] [press\_enter\_after=0|1]}: Use this to type the content into the field with id. By default, the ``Enter" key is pressed after typing unless press\_enter\_after is set to 0.
  \item \texttt{hover [id]}: Hover over an element with id.
  \item \texttt{press [key\_comb]}: Simulates the pressing of a key combination on the keyboard (e.g., Ctrl+v).
  \item \texttt{scroll [direction=down|up]}: Scroll the page up or down.
  \item \texttt{new\_tab}: Open a new, empty browser tab.
  \item \texttt{tab\_focus [tab\_index]}: Switch the browser's focus to a specific tab using its index.
  \item \texttt{close\_tab}: Close the currently active tab.
  \item \texttt{goto [url]}: Navigate to a specific URL.
  \item \texttt{go\_back}: Navigate to the previously viewed page.
  \item \texttt{go\_forward}: Navigate to the next page (if a previous `go\_back' action was performed).
  \item \texttt{stop [answer]}: Issue this action when you believe the task is complete. If the objective is to find a text-based answer, provide the answer in the bracket. If you believe the task is impossible to complete, provide the answer as ``N/A" in the bracket.
\end{itemize}

\textbf{Output Format:} 

First, generate the reasoning process for the action. Then, generate the action in the correct format. Start with a "In summary, the next action I will perform is" phrase, followed by action inside \texttt{\textasciigrave\textasciigrave\textasciigrave\textasciigrave\textasciigrave\textasciigrave}. For example, Let's think step-by-step. To add a product to the shopping cart, I need to navigate to the catalog or product section. The ``CATALOG" link is available with ID [1234]. In summary, the next action I will perform is \textasciigrave\textasciigrave\textasciigrave click [1234]\textasciigrave\textasciigrave\textasciigrave\\

\textbf{OBJECTIVE:} 

What is one of the popular search terms related to food in the image?

\textbf{PREVIOUS ACTION:} 

\textbf{IMAGES: (1)}\\

\textbf{Assistant Output:}

The task is to find a popular search term related to food. I see the search term Cereal is listed among the popular search terms, and it is related to food. So, Cereals is the answer. In summary, the next action I will perform is \texttt{\textasciigrave\textasciigrave\textasciigrave stop [Cereals]\textasciigrave\textasciigrave\textasciigrave}

\end{QAexamplebox}
\end{center}

\section{Detailed Experimental Setup}

In this section, we provide detailed descriptions of the experimental settings to facilitate reproducibility. We first introduce the baselines and benchmarks, then outline the evaluation protocol, and finally detail the hardware and training settings for \texttt{Qwen2.5-VL-7B-Instruct}~\cite{bai2025qwen2}, \texttt{InternVL2.5-8B}~\cite{chen2024expanding} and \texttt{LLaVA‑NeXT‑8B}~\cite{li2024llava}.

\subsection{Baselines and Evaluation Benchmarks}
\label{sec:baselines}

\noindent\textbf{Baselines.}
To comprehensively evaluate the effectiveness of our method, we compare against the following categories of baselines:

\textbf{Closed-source Model.}
We include GPT-4o, a strong commercial vision-language agent accessed via OpenAI's official API. All GPT-4o evaluations are conducted in a strict zero-shot setting, without additional training or in-context demonstrations. This serves as an upper bound reference.

\textbf{Open-source Agent Model.}
We evaluate against two representative open-source agents:
\begin{itemize}
    \item \textbf{Augvis}~\cite{xu2024aguvis}: a unified vision-based GUI agent that operates directly on raw screen images and performs cross-platform tasks through standardized visual interactions. Augvis is trained on a large-scale multimodal dataset with GUI grounding and reasoning annotations. It achieves strong performance on both offline and real-world benchmarks without relying on closed-source models.
    \item \textbf{ViGoRL}~\cite{sarch2025grounded}: a vision-language agent trained via reinforcement learning to perform spatially grounded reasoning. It anchors each reasoning step to specific visual coordinates and dynamically adjusts visual attention through multi-turn RL. This design enables effective fine-grained exploration and visual decision-making across complex GUI environments.
\end{itemize}

\textbf{Backbone Models.}
Our method is implemented on two strong open-source multimodal foundation models: \texttt{InternVL2.5-8B} and \texttt{Qwen2.5VL-7B-Instruct}.

\textbf{Enhanced Training or Inference Strategies.}
To further investigate the web agent capabilities of the models, we apply three widely used enhancement strategies on top of the above backbones:

\begin{itemize}
    \item \textbf{GPT Trajectory Fine-tuning}: We fine-tune the agent using successful trajectories obtained from GPT-4o on the evaluation tasks. These trajectories are collected by running GPT-4o in the test environments and extracting correct responses. As a result, this strategy introduces a powerful form of supervision, especially effective in improving performance on tasks the model has seen during fine-tuning.
    
    \item \textbf{OS-Genesis}~\cite{sun2024genesis}: This setting uses the OS-Genesis-Web dataset, which is employed to fine-tune the model. The dataset is constructed by having an agent randomly explore synthetic websites and then automatically reverse-construct corresponding tasks and aligned trajectories. The dataset contains high-quality expert demonstrations across various domains (e.g. shopping, reddit and classifieds). 
    
    \item \textbf{Tree Search Inference}~\cite{koh2024tree}: A manually crafted inference pipeline that enables language model agents to perform multi-step planning in interactive web environments. To ensure experimental efficiency, we adopt the following setup: the agent generates five candidate actions at each step and selects two valid ones for simulation. The resulting states are evaluated using the model itself as a value function, and only high-scoring paths are retained while low-value branches are pruned. The search is performed with a depth of three steps. This method supports strategic exploration and multi-step reasoning during inference without requiring any additional training.
\end{itemize}

\textbf{SCALE.} We leverage the SCALE framework to collect trajectory data through autonomous interaction with corresponding web environments. It enables agents to explore, assess, and adapt beyond their cognitive boundaries in dynamic and unfamiliar web settings. Based on these, we use another model to construct structured tasks (e.g., single-step, multi-step, QA), which are then used to fine-tune the base model. This enables the agent to improve through its own experience without relying on human annotations.

\textbf{SCALE-20k.}
To further evaluate the effectiveness of our collected SCALE-20k dataset, we fine-tune another model, \texttt{LLaVA‑NeXT‑8B}~\cite{li2024llava}, which has a different architecture, directly on the data collected by \texttt{Qwen2.5-VL-7B-Instruct} and \texttt{InternVL2.5-8B}, and compare its performance with its own baseline.

\noindent\textbf{Evaluation Benchmarks.}
We evaluate our method and baselines on two established benchmarks: VisualWebArena~\cite{koh2024visualwebarena} and WebVoyager~\cite{he2024webvoyager}, which together cover both predefined and real-world web environments.

\begin{itemize}
\item \textbf{VisualWebArena:}
VisualWebArena includes evaluation tasks from three representative websites, Shopping, Classifieds, and Reddit, along with a small number of cross-domain tasks. For each individual website, we exclusively train the agent on the data collected by SCALE within that website's environment, ensuring that evaluation remains domain-specific. To account for varying website complexity, we set the maximum exploration depth to \textbf{60-25}, where the agent runs up to 60 self-update rounds with 25 interaction steps each. Since over-exploration may introduce noisy actions, we evaluate models trained with different exploration depths and report results from the depth with the highest validation success rate.

\item \textbf{WebVoyager:}
WebVoyager is a large-scale benchmark designed for evaluating web agents in real-world websites. In our experiments, we evaluate on all 15 websites in WebVoyager, using the first 11 tasks from each site to construct a comprehensive evaluation suite. For data collection, SCALE is used to explore each site at depths 10-25, where the agent runs up to 10 self-update rounds with 25 interaction steps each. We then combine exploration data from all 15 websites to train a general-purpose agent for evaluation. Due to the lack of official GPT-4o trajectories and the limited generalizability of Tree Search-style methods to highly dynamic and diverse real-world websites, we exclude those two baselines from this benchmark.
\end{itemize}

\textbf{Input Standardization:}
To ensure consistency across all methods and backbones, we standardize the input format for all agents. Specifically, we remove all auxiliary metadata (e.g., accessibility trees or HTML DOM representations) provided by the original benchmarks and retain only the visual screenshot and its corresponding set-of-mark (SOM)~\cite{yang2023set,wan2024omniparser}. For different model backbones, image resolution is automatically adjusted to meet the input size constraints, ensuring stable inference across evaluation settings.

\subsection{Training Hardware and Hyperparameters}
\label{sec:hardware}

\noindent\textbf{Hardware.}
During the exploration phase, we used a total of 8 GPUs for SCALE deployment, model training, and updates. All training and evaluation experiments are conducted on NVIDIA RTX 4090 GPUs. We utilize 2 or 4 GPUs for model fine-tuning and inference.

\noindent\textbf{Training Hyperparameters.}
We fine-tune three different vision-language models in our experiments: \texttt{Qwen2.5-VL-7B-Instruct}, \texttt{InternVL2.5-8B} and \texttt{LLaVA‑NeXT‑8B}.\\

\textbf{Qwen2.5-VL-7B-Instruct:} We fine-tune the model using the \texttt{LLaMA Factory} framework with LoRA-based supervised fine-tuning. The main configurations are as follows:
\begin{itemize}
\item \textbf{LoRA:} Applied to all modules with rank 8.
\item \textbf{Input:} Images resized to a maximum of 921,600 pixels; token cutoff length set to 4096.
\item \textbf{Batching:} 2 GPUs, with per-device batch size 1 and gradient accumulation of 8.
\item \textbf{Optimization:} Learning rate 1e-4, cosine decay, warmup ratio 3\%, 15 training epochs. Select approximately 1000 steps checkpoint for testing (the checkpoint that performed best during training).
\item \textbf{Precision:} bf16 and FlashAttention v2 enabled.\\
\end{itemize}

\textbf{InternVL2.5-8B:} We follow the official LoRA training pipeline with the following key hyperparameters:
\begin{itemize}
\item \textbf{LoRA:} Only LLM is fine-tuned with LoRA (rank 8); vision backbone, MLP, and LLM weights are frozen.
\item \textbf{Input:} Dynamic image resolution with thumbnails enabled; forced image size set to 448; max dynamic patch is set to 4.
\item \textbf{Batching:} 2 GPUs, with per-device batch size 1 and gradient accumulation of 8.
\item \textbf{Optimization:} Learning rate 1e-4, cosine schedule, weight decay 0.05, warmup ratio 3\%, 10 training epochs. 
\item \textbf{Precision:} bf16 training with gradient checkpointing enabled.\\
\end{itemize}

\textbf{LLaVA‑NeXT‑8B:} We fine-tune the model using the \texttt{LLaMA Factory} framework with LoRA-based supervised fine-tuning. The main configurations are as follows:
\begin{itemize}
\item \textbf{LoRA:} Applied to all modules with rank 8.
\item \textbf{Input:} Images resized to a maximum of 921,600 pixels; token cutoff length set to 4096.
\item \textbf{Batching:} 4 GPUs, with per-device batch size 1 and gradient accumulation of 8.
\item \textbf{Optimization:} Learning rate 1e-4, cosine decay, warmup ratio 3\%, 15 training epochs. Select approximately 500 steps checkpoint for testing (the checkpoint that performed best during training).
\item \textbf{Precision:} bf16 and FlashAttention v2 enabled.
\end{itemize}

\section{Quality Evaluation of SCALE-20k}
In this section, we first introduce three metrics for evaluating dataset quality and provide corresponding evaluation prompts.

\subsection{Evaluation Metrics}
\label{sec:evaluation metrics}
To evaluate the diversity and quality of the \textbf{SCALE-20k} dataset, we adopt three complementary metrics:

\textbf{Task Diversity.} We follow the procedure described in the main paper and embed each task description using Sentence-BERT~\cite{reimers-gurevych-2019-sentence}. We then compute the average pairwise cosine distance between all task embeddings. A larger average distance indicates greater semantic variation across tasks, reflecting broader coverage of task types.

\textbf{Reasoning Quality.} 
To assess the quality of the agent’s reasoning, we employ GPT-4o as an automatic evaluator. For each task instance, the evaluator is given the task objective, the generated reasoning, the proposed next action, and screenshots for context. GPT-4o scores the reasoning on a 1-10 scale by examining whether it is accurate, concise when appropriate, free of hallucinations, and tightly aligned with the proposed action and task. A brief justification is produced together with the numerical score. The final reasoning quality score is obtained by averaging the GPT-4o scores across all instances and linearly mapping the result to the 0-1 range.

\textbf{Answer Quality.}
Similarly, to evaluate the correctness and appropriateness of the agent’s proposed next action (or final answer), we employ GPT-4o as an automatic evaluator. For each task instance, the evaluator is given the task instruction, the assistant’s reasoning and action, and screenshots. GPT-4o scores the answer on a 1-10 scale by examining whether the action is valid for the current observation, follows all agent side constraints (e.g., one action at a time, whitelist-restricted action types, and required output format), and constitutes the best or near-best step toward accomplishing the objective. A brief justification is produced together with the numerical score. The final answer quality score is obtained by averaging the GPT-4o scores across all instances and linearly mapping the result to the 0-1 range.

In summary, these three metrics jointly capture the semantic diversity of tasks and the quality of reasoning and actions, providing a comprehensive assessment of the SCALE-20k dataset.

\subsection{Prompts for Quality Evaluation}
\label{sec:evaluation prompts}
The exact evaluation prompts used for reasoning quality and answer quality are provided below.

\noindent\textbf{Reasoning Quality Evaluation Prompt}
\newtcolorbox{reasonpromptbox}[1][]{
  colback=gray!10!white,
  colframe=black,
  width=\columnwidth,
  arc=2mm,
  boxrule=0.5pt,
  title=\textbf{Reasoning Quality Evaluation Prompt},
  fonttitle=\normalsize\bfseries,
  enhanced,
  breakable,
  sharp corners,
  #1
}

\begin{reasonpromptbox}

You are a professional web-agent evaluator focusing ONLY on the agent's reasoning quality.

You will be given:
\begin{enumerate}[leftmargin=12pt, label=\arabic*)]
    \item OBJECTIVE | the task goal.
    \item Agent's reasoning and proposed next action (assistant content).
    \item Screenshots for context.
\end{enumerate}

\textbf{Your Objective:}
\begin{itemize}[leftmargin=12pt, label=-]
    \item Evaluate ONLY the REASONING (not the action's optimality). Prioritize ACCURACY over length.
    \item If the task is simple, concise reasoning is preferred; if complex, more elaboration is acceptable.
    \item The reasoning MUST be tightly aligned with the final action/answer: no off-topic chains, and no mismatch between reasoning and the proposed action.
    \item Check compliance with agent reasoning rules (one action at a time, correct format reasoning-to-action flow), but do NOT score the action optimality here.
\end{itemize}

\textbf{Evaluation Scale (1--10):}
\begin{itemize}[leftmargin=12pt, label=-]
    \item \textbf{9--10:} Clear, accurate, stepwise when needed; tightly linked to the final action; no hallucinations; reasoning references the page context appropriately; concise when the task is simple.
    \item \textbf{7--8:} Mostly clear and accurate with minor omissions; still aligned with the final action; minor verbosity or slight gaps.
    \item \textbf{5--6:} Partial accuracy; noticeable gaps, vague steps, or weak linkage to the final action; occasional rule looseness.
    \item \textbf{3--4:} Several logical issues; unclear or verbose without need; weak/contradictory linkage to the action.
    \item \textbf{1--2:} Largely incoherent or irrelevant; contradicted by the final action; ignores rules or context.
\end{itemize}

\textbf{Guidance:}
\begin{itemize}[leftmargin=12pt, label=-]
    \item Penalize verbosity without substance, hallucinations, or reasoning–action inconsistency.
    \item Reward concise-but-correct reasoning for simple objectives; reward stepwise clarity for complex ones.
    \item Do not evaluate whether the chosen action is the best one (that is scored separately).
\end{itemize}

\textbf{Output Format}

Provide a brief justification (1–3 sentences) about reasoning quality, then ONLY this JSON:
\\\{
\\``reason": ``\textless{}brief justification about reasoning quality\textgreater{}",
\\``score": X
\\\}

\textbf{OBJECTIVE:}\\
\{objective text\}

\textbf{Agent's reasoning and action:}\\
\{assistant text\}

\end{reasonpromptbox}

\noindent\textbf{Answer Quality Evaluation Prompt}
\newtcolorbox{anssonpromptbox}[1][]{
  colback=gray!10!white,
  colframe=black,
  width=\columnwidth,
  arc=2mm,
  boxrule=0.5pt,
  title=\textbf{Answer Quality Evaluation Prompt},
  fonttitle=\normalsize\bfseries,
  enhanced,
  breakable,
  sharp corners,
  #1
}

\begin{anssonpromptbox}
You are a professional web-agent evaluator focusing ONLY on the action/answer quality w.r.t. the OBJECTIVE.

You will be given:
\begin{enumerate}[leftmargin=12pt, label=\arabic*)]
    \item OBJECTIVE | the task goal.
    \item Agent's reasoning and proposed next action (assistant content).
    \item Screenshots for context.
\end{enumerate}

Your Objective:\\
- Evaluate whether the proposed NEXT ACTION (or final answer) is the BEST choice for the current environment/state.\\
- STRICTLY check the following RULES are obeyed:
\begin{enumerate}[label=\arabic*., leftmargin=2em]
    \item The action must be VALID given the current observation.
    \item Only ONE action at a time.
    \item Follow examples to reason step by step and then issue the next action.
    \item Correct output format: must start with the phrase: ``In summary, the next action I will perform is" followed by the action inside triple backticks.
\end{enumerate}
- The action MUST be one of the ALLOWED ACTIONS (whitelist):\\
\hspace*{2em}\textbf{Page Operation Actions}:
\begin{enumerate}[label=-, leftmargin=4em]
    \item click [id]
    \item type [id] [content]   (optional enter suppression: type [id] [content] [0])
    \item hover [id]
    \item press [key comb]
    \item scroll [down] \textbar~scroll [up]
\end{enumerate}
\hspace*{2em}\textbf{Tab Management Actions}:
\begin{enumerate}[label=-, leftmargin=4em]
    \item new\_tab
    \item tab\_focus [tabindex]
    \item close\_tab
\end{enumerate}
\hspace*{2em}\textbf{URL Navigation Actions}:
\begin{enumerate}[label=-, leftmargin=4em]
    \item goto [url]
    \item go\_back
    \item go\_forward
\end{enumerate}
\hspace*{2em}\textbf{Completion Action}:
\begin{enumerate}[label=-, leftmargin=4em]
    \item stop [answer]
\end{enumerate}

\textbf{Evaluation Scale (1--10):}
\begin{itemize}[leftmargin=12pt, label=-]
    \item \textbf{9--10:} The action is clearly optimal/near-optimal for the current state; valid, correctly formatted, follows all rules; directly progresses or completes the OBJECTIVE without harmful side effects.
    \item \textbf{7--8:} Good choice that likely helps progress; minor inefficiency or small risk, but still valid and compliant.
    \item \textbf{5--6:} Partially helpful or somewhat redundant; may miss a clearer step or have limited progress value.
    \item \textbf{3--4:} Poor choice; low impact, potentially invalid for the state, or weak alignment with the objective; may break minor format/rule expectations.
    \item \textbf{1--2:} Clearly wrong or counterproductive; violates core rules/format, not in the allowed action set, or does not move toward the objective.
\end{itemize}

\textbf{Guidance:}
\begin{itemize}[leftmargin=12pt, label=-]
    \item Penalize any rule or format violation, wrong action type, multiple actions at once, or inapplicable id.
    \item Reward the most effective, necessary, and safe action that best advances/completes the objective now.
    \item Do NOT score the reasoning clarity here (that is evaluated separately).
\end{itemize}

\textbf{Output Format}
Provide a brief justification (1–3 sentences) about action optimality and rule compliance, then ONLY this JSON:
\\\{
\\``reason": ``\textless{}brief justification about action/answer quality\textgreater{}",
\\``score": X
\\\}

\textbf{OBJECTIVE:}\\
\{objective text\}

\textbf{Agent's reasoning and action:}\\
\{assistant text\}
\end{anssonpromptbox}



\end{document}